\documentclass[twoside]{article}

\usepackage[accepted]{aistats2022}

\usepackage{microtype}
\usepackage{graphicx}
\usepackage{subcaption}
\usepackage{booktabs} 
\usepackage{hyperref}

\usepackage[round]{natbib}

\usepackage{amsmath,amsfonts,bm}

\def\eqref#1{equation~\ref{#1}}

\def\1{\bm{1}}

\newcommand{\reals}{\mathbb R}

\def\rx{{\textnormal{x}}}
\def\ry{{\textnormal{y}}}
\def\rz{{\textnormal{z}}}

\DeclareMathAlphabet{\mathsfit}{\encodingdefault}{\sfdefault}{m}{sl}
\SetMathAlphabet{\mathsfit}{bold}{\encodingdefault}{\sfdefault}{bx}{n}

\newcommand{\maximize}{\operatorname*{maximize}}

\usepackage[utf8]{inputenc} 
\usepackage[T1]{fontenc}    
\usepackage{url}            
\usepackage{amsfonts}       
\usepackage{nicefrac}       
\usepackage{adjustbox}
\usepackage{anonymized}
\usepackage{amssymb,amsmath,amsthm}
\usepackage{tikz}
\usepackage{enumitem}
\usepackage{verbatim}
\usepackage{multirow}
\usepackage{algorithm,algorithmic}
\usepackage[rightcaption]{sidecap}
\usepackage{wrapfig}
\usetikzlibrary{positioning,decorations.pathreplacing,decorations.markings,calc,arrows,hobby,backgrounds,patterns}
\newcommand{\hobbyconvexpath}[2]{
[   
    create hobbyhullnodes/.code={
        \global\edef\namelist{#1}
        \foreach [count=\counter] \nodename in \namelist {
            \global\edef\numberofnodes{\counter}
            \node at (\nodename)
[draw=none,name=hobbyhullnode\counter] {};
        }
        \node at (hobbyhullnode\numberofnodes)
[name=hobbyhullnode0,draw=none] {};
        \pgfmathtruncatemacro\lastnumber{\numberofnodes+1}
        \node at (hobbyhullnode1)
[name=hobbyhullnode\lastnumber,draw=none] {};
    },
    create hobbyhullnodes
]
($(hobbyhullnode1)!#2!-90:(hobbyhullnode0)$)
\pgfextra{
  \gdef\hullpath{}
\foreach [
    evaluate=\currentnode as \previousnode using int(\currentnode-1),
    evaluate=\currentnode as \nextnode using int(\currentnode+1)
    ] \currentnode in {1,...,\numberofnodes} {
    \ifnum\currentnode=1\relax
    \xdef\hullpath{([closed=true]$(hobbyhullnode\currentnode)!#2!180:(hobbyhullnode\previousnode)$)
  ..($(hobbyhullnode\nextnode)!0.5!(hobbyhullnode\currentnode)$)}
    \else
    \xdef\hullpath{\hullpath
  ..($(hobbyhullnode\currentnode)!#2!180:(hobbyhullnode\previousnode)$)
  ..($(hobbyhullnode\nextnode)!0.5!(hobbyhullnode\currentnode)$)}
    \fi
    \ifx\currentnode\numberofnodes
    \else
    \xdef\hullpath{\hullpath
  ..($(hobbyhullnode\nextnode)!#2!-90:(hobbyhullnode\currentnode)$)}
    \fi
}
}
\hullpath
}

\newlength{\hatchspread}
\newlength{\hatchthickness}
\newlength{\hatchshift}
\newcommand{\hatchcolor}{}
\tikzset{hatchspread/.code={\setlength{\hatchspread}{#1}},
         hatchthickness/.code={\setlength{\hatchthickness}{#1}},
         hatchshift/.code={\setlength{\hatchshift}{#1}},
         hatchcolor/.code={\renewcommand{\hatchcolor}{#1}}}
\tikzset{hatchspread=3pt,
         hatchthickness=0.07pt,
         hatchshift=1pt,
         hatchcolor=gray}
\pgfdeclarepatternformonly[\hatchspread,\hatchthickness,\hatchshift,\hatchcolor]
   {custom north west lines}
   {\pgfqpoint{\dimexpr-2\hatchthickness}{\dimexpr-2\hatchthickness}}
   {\pgfqpoint{\dimexpr\hatchspread+2\hatchthickness}{\dimexpr\hatchspread+2\hatchthickness}}
   {\pgfqpoint{\dimexpr\hatchspread}{\dimexpr\hatchspread}}
   {
    \pgfsetlinewidth{\hatchthickness}
    \pgfpathmoveto{\pgfqpoint{0pt}{\dimexpr\hatchspread+\hatchshift}}
    \pgfpathlineto{\pgfqpoint{\dimexpr\hatchspread+0.15pt+\hatchshift}{-0.15pt}}
    \ifdim \hatchshift > 0pt
      \pgfpathmoveto{\pgfqpoint{0pt}{\hatchshift}}
      \pgfpathlineto{\pgfqpoint{\dimexpr0.15pt+\hatchshift}{-0.15pt}}
    \fi
    \pgfsetstrokecolor{\hatchcolor}
    \pgfusepath{stroke}
   }
\pgfdeclarepatternformonly[\hatchspread,\hatchthickness,\hatchshift,\hatchcolor]
   {custom north east lines}
   {\pgfqpoint{\dimexpr-2\hatchthickness}{\dimexpr-2\hatchthickness}}
   {\pgfqpoint{\dimexpr\hatchspread+2\hatchthickness}{\dimexpr\hatchspread+2\hatchthickness}}
   {\pgfqpoint{\dimexpr\hatchspread}{\dimexpr\hatchspread}}
   {
    \pgfsetlinewidth{\hatchthickness}
    \pgfpathmoveto{\pgfqpoint{\dimexpr\hatchshift-0.15pt}{-0.15pt}}
    \pgfpathlineto{\pgfqpoint{\dimexpr\hatchspread+0.15pt}{\dimexpr\hatchspread-\hatchshift+0.15pt}}
    \ifdim \hatchshift > 0pt
      \pgfpathmoveto{\pgfqpoint{-0.15pt}{\dimexpr\hatchspread-\hatchshift-0.15pt}}
      \pgfpathlineto{\pgfqpoint{\dimexpr\hatchshift+0.15pt}{\dimexpr\hatchspread+0.15pt}}
    \fi
    \pgfsetstrokecolor{\hatchcolor}
    \pgfusepath{stroke}
   }
\usepgfplotslibrary{fillbetween}
\usepackage{cancel}
\usepackage{transparent}

\newtheorem{lemma}{Lemma} 

\newtheorem{definition}{Definition}
\newtheorem{proposition}{Proposition}

\begin{document}

\twocolumn[

\aistatstitle{Optimal channel selection with discrete QCQP}

\aistatsauthor{Yeonwoo Jeong$^{*1}$ \And Deokjae Lee$^{*1}$ \And  Gaon An $^{1}$\And Changyong Son $^{2}$\And Hyun Oh Song $^{\dagger 1}$}
\aistatsaddress{ $^1$ Department of Computer Science and Engineering, Seoul National University \\$^2$Samsung Advanced Institute of Technology}]

\begin{abstract}
	 Reducing the high computational cost of large convolutional neural networks is crucial when deploying the networks to resource-constrained environments. We first show the greedy approach of recent channel pruning methods ignores the inherent quadratic coupling between channels in the neighboring layers and cannot safely remove inactive weights during the pruning procedure. Furthermore, due to these inactive weights, the greedy methods cannot guarantee to satisfy the given resource constraints and deviate with the true objective. In this regard, we propose a novel channel selection method that optimally selects channels via discrete QCQP, which provably prevents any inactive weights and guarantees to meet the resource constraints \emph{tightly} in terms of FLOPs, memory usage, and network size. We also propose a quadratic model that accurately estimates the \emph{actual inference time} of the pruned network, which allows us to adopt inference time as a resource constraint option. Furthermore, we generalize our method to extend the selection granularity beyond channels and handle non-sequential connections. Our experiments on CIFAR-10 and ImageNet show our proposed pruning method outperforms other fixed-importance channel pruning methods on various network architectures.  
\end{abstract}

\section{Introduction}
Deep neural networks are the bedrock of artificial intelligence tasks such as object detection, speech recognition, and natural language processing \citep{detection, speech, nlp}. While modern networks have hundreds of millions to billions of parameters to train, recent works show that these parameters are highly redundant and can be pruned without significant loss in accuracy \citep{han2015learning, guo2016dynamic}. This discovery has led practitioners to desire training and running the models on resource-constrained mobile devices, provoking a large body of research on network pruning. 

Unstructured pruning, however, does not directly lead to any practical acceleration or memory footprint reduction due to poor data locality \citep{wen2016learning}, and this motivated research on structured pruning to achieve practical usage under limited resource budgets. To this end, a line of research on channel pruning considers completely pruning the convolution filters along the input and output channel dimensions, where the resulting pruned model becomes a smaller dense network suited for practical acceleration and memory footprint reduction \citep{pfec, luo2017thinet, fpgm, wen2016learning, sfp}. 

While many channel pruning methods select channels greedily, which is easy to model and optimize, they cannot safely remove inactive weights during the pruning procedure. As a result, these greedy approaches suffer from discrepancies with the true objective and cannot strictly satisfy the required resource constraints during the pruning process. 

The ability to specify hard target resource constraints into the pruning optimization process is important since this allows the user to run the pruning and optional finetuning process only once. When the pruning process ignores the target specifications, the users may need to apply multiple rounds of pruning and finetuning until the specifications are eventually met, resulting in an extra computation overhead \citep{han2015learning, sfp, liu2017learning}. 

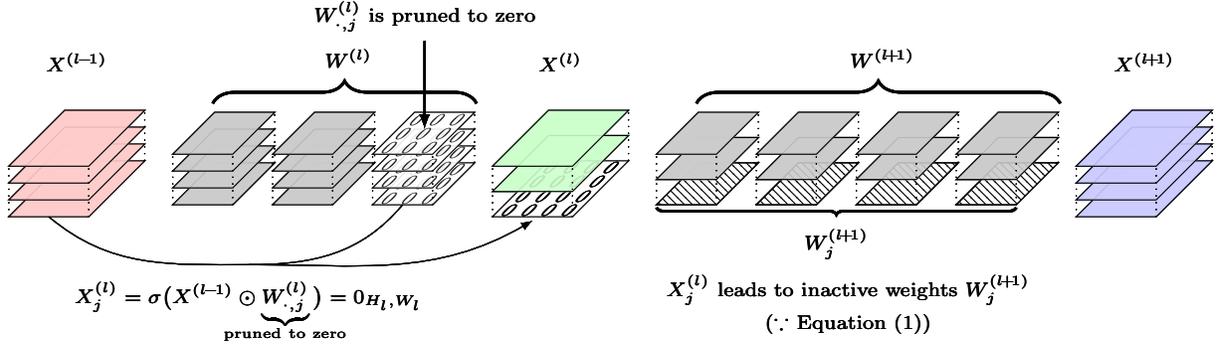
\begin{figure*}[ht]
    \centering
\def\rot{0}
\def\dd{0.3}
\def\ddd{0.2}
\def\channelwidth{6}
\resizebox{2.0\columnwidth}{!}{
\pgfdeclarelayer{fig}\pgfdeclarelayer{bg}
\pgfsetlayers{fig,bg}  
\begin{tikzpicture}[decoration={brace}][scale=1,every node/.style={minimum size=1cm},on grid]
\foreach \num in {0,1,2,3}{
    \begin{pgfonlayer}{bg}

    \end{pgfonlayer}
    \begin{pgfonlayer}{fig}
    \foreach \i in {0,1,2,3} {
        \begin{scope}[  
            yshift={\channelwidth*\i},every node/.append style={
                 yslant=0,xslant=1,rotate=\rot},yslant=0,xslant=1,rotate=\rot
              ]              
              \coordinate (A1\i) at (0,0);
              \coordinate (B1\i) at (1.9,0.15);
              \coordinate (B2\i) at (3.15,0.15);
              \coordinate (B3\i) at (4.4,0.15);
              \coordinate (A3\i) at (13.35,0);
              \draw[fill,red!20,opacity=0.9] (A1\i) rectangle ++(1,0.7);
              \draw (A1\i) rectangle ++(1,0.7);
              \draw[fill,gray!40,opacity=0.9] (B1\i) rectangle ++(0.75,0.525);
              \draw (B1\i) rectangle ++(0.75,0.525);
              \draw[fill,gray!40,opacity=0.9] (B2\i) rectangle ++(0.75,0.525);
              \draw (B2\i) rectangle ++(0.75,0.525);
              \draw[fill,white,opacity=0.9] (B3\i) rectangle ++(0.75,0.525);
              \draw (B3\i) rectangle ++(0.75,0.525);
              \draw[fill,blue!20,opacity=0.9] (A3\i) rectangle ++(1,0.7);
              \draw (A3\i) rectangle ++(1,0.7);
              
              \foreach\n in {1,3,5}{
              \foreach\m in {1,3,5}{
              \node () at ($(B3\i)+(\n*0.125,\m*0.0875)$) {\tiny $0$};
              }
              }
        \end{scope}
        }
        \foreach \i in {0,1,2} {
        \begin{scope}[  
            yshift={1.5*\channelwidth*\i},every node/.append style={
                 yslant=0,xslant=1,rotate=\rot},yslant=0,xslant=1,rotate=\rot
              ]              
              \coordinate (A2\i) at (6.05,0);
              \coordinate (B4\i) at (7.95,0.15);
              \coordinate (B5\i) at (9.2,0.15);
              \coordinate (B6\i) at (10.45,0.15);
              \coordinate (B7\i) at (11.7,0.15);
              \ifnum\i=0
              \def\c{white}
              \def\d{white}
              \else 
              \def\c{gray!40}
              \def\d{green!20}
              \fi
              \draw[fill,\d,opacity=0.9] (A2\i) rectangle ++(1,0.7);
              \draw (A2\i) rectangle ++(1,0.7);
              \draw[fill,\c,opacity=0.9] (B4\i) rectangle ++(0.75,0.525);
              \foreach\j in {4,5,6,7}{
              \draw[fill,\c,opacity=0.9] (B\j\i) rectangle ++(0.75,0.525);
              \draw (B\j\i) rectangle ++(0.75,0.525);
              \ifnum\i=0
              \foreach\n in {1,3,5,7}{
              \foreach\m in {1,3,5,7}{
              \node () at ($(A20)+(\n*0.125,\m*0.0875)$) {\tiny $0$};
              }
              }
              \draw[pattern=north west lines, pattern color = black] (B\j\i) rectangle ++(0.75,0.525);
              \fi
              }
        \end{scope}
        }
    \node () at ($(A13)+(0.85,1.3)$) {\scriptsize $X^{(l\!-\!1)}$};   
    \node () at ($(A22)+(0.85,1.3)$) {\scriptsize $X^{(l)}$};    
    \node () at ($(A33)+(0.85,1.3)$) {\scriptsize $X^{(l\!+\!1)}$};
    
    \end{pgfonlayer}
    \begin{pgfonlayer}{bg}
    \begin{scope}[  
            yslant=0,xslant=1,rotate=\rot
              ]
    \foreach\j in {1,2,3}{
    \ifnum\j=2
    \def\k{2}
    \else
    \def\k{3}
    \fi
    \draw[dotted] (A\j0) -- (A\j\k);
    \draw[dotted] ($(A\j0) + (1,0)$) -- ($(A\j\k) + (1,0)$);
    \draw[dotted] ($(A\j0) + (1,0.7)$) -- ($(A\j\k) + (1,0.7)$);
    }
    \foreach\j in {1,2,3,4,5,6,7}{
    \ifnum\j<4
    \def\k{3}
    \else
    \def\k{2}
    \fi
    \draw[dotted] (B\j0) -- (B\j\k);
    \draw[dotted] ($(B\j0) + (0.75,0)$) -- ($(B\j\k) + (0.75,0)$);
    \draw[dotted] ($(B\j0) + (0.75,0.525)$) -- ($(B\j\k) + (0.75,0.525)$);
    }
    \end{scope}
    \draw[thick,decorate,decoration={brace,amplitude=8pt,raise=1pt},line width=1pt] ($(B13)+(0.5,0.6)$) -- ($(B33)+(1.3,0.6)$);
    \node () at ($(B13)!0.5!(B33) + (0.95,1.2)$) {\scriptsize $W^{(l)}$};
    \draw[thick,decorate,decoration={brace,amplitude=8pt,raise=1pt},line width=1pt] ($(B42)+(0.5,0.6)$) -- ($(B72)+(1.3,0.6)$);
    \node () at ($(B42)!0.5!(B72) + (0.95,1.2)$) {\scriptsize $W^{(l\!+\!1)}$};
    \coordinate (o) at (3,-0.6);
    \coordinate (p) at (5.2,2.2);
    \draw[-latex,thick] (p) -- ($(p)+(0,-1.1)$);
    \node[above] (pp) at (p) {\scriptsize $W^{(l)}_{\cdot,j}$ is pruned to zero};
    \draw ($(A10)+(0.5,0)$) to[out=-40,in=180] (o);
    \draw ($(B30)+(0.375,0)$) to[out=230,in=0] (o);
    \draw[-latex] (o) to[out=0,in=210] ($(A20)+(0.5,-0.05)$);
    \node[align=center] (l) at (3,-1.2) {\scriptsize $X^{(l)}_j=\sigma\big(X^{(l\!-\!1)}\odot \underbrace{W^{(l)}_{\cdot,j}}_{\text{\makebox[0pt]{pruned to zero}}}\big)=0_{H_l, W_l} $};
    \node[align=center] (l) at (10.5,-1.1) {\scriptsize $X^{(l)}_j$ leads to inactive weights $W^{(l\!+\!1)}_{j}$ \\ \scriptsize ($\because$ \Cref{eq:toyinactive})};
    \node[below,yshift=-5] () at ($($(B70)+(0.75,0)$)!0.5!(B40)$) {\scriptsize $W^{(l\!+\!1)}_{j}$};
    \end{pgfonlayer}
    \begin{pgfonlayer}{fig}
    \begin{scope}[  
            yshift={0},every node/.append style={
                 yslant=0,xslant=1,rotate=\rot},yslant=0,xslant=1,rotate=\rot
        ]
        \draw[thin,decorate,decoration={brace,raise=1pt},line width=1pt] ($(B70)+(0.75,0)$) -- (B40) ;
              
    \end{scope}
    \end{pgfonlayer}
}
\end{tikzpicture}
}
\caption{Illustration of a channel pruning procedure that leads to inactive weights. When $j$-th output channel of $l$-th convolution weights  $W^{(l)}_{\cdot,j}$ is pruned, \ie\  $W^{(l)}_{\cdot,j}=0_{C_{l\!-\!1},K_l,K_l}$, then the $j$-th feature map of $l$-th layer $X^{(l)}_j$ should also be $0$. Consequently, $X^{(l)}_j$ yields inactive weights $W^{(l+1)}_{j}$. Note that we use $W^{(l)}_{\cdot,j}$ to denote the tensor $W^{(l)}_{\cdot,j,\cdot,\cdot}$, following the indexing rules of NumPy \citep{van2011numpy}.} 
\label{fig:fig1}
\end{figure*}

Our contributions can be summarized as follows:
\begin{itemize}\itemsep=1pt
	 \item We propose an optimal channel selection method which satisfies the user-specified constraints \textit{tightly} in terms of FLOPs, memory usage, and network size, and directly maximizes the importance of neurons in the pruned network. 
    \item We propose a new quadratic model that accurately estimates the \textit{inference time} of a pruned network \textit{without direct deployment}.
    \item We extend our method to increase the pruning granularity beyond channels and simultaneously prune channels and spatial patterns in the individual 2D convolution filters.
    \item We generalize our method to handle nonsequential connections (skip additions and skip concatenations). 
\end{itemize}

Our experiments on CIFAR-10 and ImageNet datasets show the state of the art results compared to other fixed-importance channel pruning methods.


\section{Motivation}
In this section, we first discuss the motivation of our method more concretely. The weights of a sequential CNN can be expressed as a sequence of 4-D tensors, $W^{(l)} \in \reals^{C_{l\!-\!1}\times C_l \times K_l \times K_l} ~~\forall l\in [L]$, where $C_{l\!-\!1}$, $C_l$, and $K_l$ represent the number of input channels, the number of output channels, and the filter size of the $l$-th convolution weight tensor, respectively. We denote the feature map after the $l$-th convolution as $X^{(l)}\in \reals^{C_l \times H_l \times W_l}$. Concretely, $X_j^{(l)} = \sigma( X^{(l\!-\!1)} \odot W_{\cdot,j}^{(l)} )= \sigma( \sum_{i=1}^{C_{l\!-\!1}} X_i^{(l\!-\!1)} \ast W_{i,j}^{(l)} )$ for $j \in [C_l]$, where $\sigma$ is the activation function, $\ast$ denotes 2-D convolution operation, and $\odot$ denotes the sum of channel-wise 2-D convolutions. Now consider pruning these weights in channel-wise direction. We show that with naive channel-wise pruning methods, we cannot exactly specify the target resource constraints due to unpruned inactive weights and deviate away from the true objective by ignoring quadratic coupling between channels in the neighboring layers.

\subsection{Inactive weights}
According to \citet{han2015learning}, network pruning produces dead neurons with zero input or output connections. These dead neurons cause \emph{inactive} weights\footnote{Rigorous mathematical definition of inactive weights is provided in Supplementary material  C.},  which do not affect the final output activations of the pruned network. These inactive weights are not excluded automatically through the standard pruning procedure and require additional post-processing which relies on ad-hoc heuristics. For example, \Cref{fig:fig1} shows a standard channel pruning procedure that deletes weights across the output channel direction but fails to prune the inactive weights. Concretely, deletion of weights on $j$-th output channel of $l$-th convolution layer leads to $W^{(l)}_{\cdot,j}=0_{C_{l\!-\!1},K_l,K_l}$. Then, $X^{(l)}_j$ becomes a dead neuron since $X^{(l)}_j = \sigma(X^{(l\!-\!1)} \odot W^{(l)}_{\cdot,j})  = \sigma(\sum_{i=1}^{C_{l\!-\!1}} X^{(l\!-\!1)}_i \ast W^{(l)}_{i,j}) = 0_{H_l,W_l}$.

The convolution operation on the dead neuron results in a trivially zero output as below:
\footnotesize
\begin{align}
    X^{(l\!+\!1)}_p &= \sigma\left(\sum_{i=1}^{C_{l}} X^{(l)}_i \ast W^{(l\!+\!1)}_{i,p}\right)\nonumber \\
    &= \sigma\Bigg(\sum_{i=1}^{C_{l}} \mathds{1}_{i\neq j} X^{(l)}_i \ast W^{(l\!+\!1)}_{i,p} + \underbrace{\underbrace{X^{(l)}_j}_{\text{dead}} \ast \underbrace{W^{(l\!+\!1)}_{j,p}}_{\text{inactive}}}_{=0_{H_{l\!+\!1},W_{l\!+\!1}}} \Bigg).
    \label{eq:toyinactive}
\end{align}
\normalsize
\Cref{eq:toyinactive} shows that the dead neuron $X^{(l)}_j$ causes weights $W^{(l\!+\!1)}_{j,p}, \forall p\in [C_{l\!+\!1}]$ to be inactive. Such inactive weights do not account for the actual resource usage, even when they remain in the pruned network, which prevents the exact modeling of the user-specified resource constraints (FLOPs, memory usage, or network size). Furthermore, inactive weights unpruned during the pruning procedure becomes a bigger problem for nonsequential convolutional networks due to their skip connections, which is discussed in \Cref{subsec:nonseq}. To address thess problems, we introduce a quadratic optimization-based algorithm that provably eliminates all the inactive weights during the pruning procedure.

\input{fig/toy_ex_v2.tex}

\subsection{Quadratic coupling}
\label{subsec:quad} 

Most of the existing channel pruning methods select channels with regard to their importance. However, measuring a channel's contribution to the network should also take into account the channels in the neighboring layers, as illustrated in \Cref{fig:toy_ex}. In the example, we define the importance of a channel as the absolute sum of the weights in the channel, as in \citet{pfec}, and assume the objective is to maximize the absolute sum of the weights in the whole pruned network, excluding the inactive weights. We compare two different channel selection methods: (a) a standard greedy channel selection method which greedily selects each channel in layerwise manner, and (b) our selection method that optimally considers the effect of the channels in neighboring layers, which will be described in \Cref{sec:method}.  As a result of running each selection algorithms, (a) will prune the second output channel of the first convolution and the third output channel of the second convolution, and (b) will prune the first output channel of the first convolution, the third output channel of the second convolution, and the first input channel of the second convolution. The objective values for each pruned networks are (a) $18$  and (b) $21$, respectively. 

This shows that the coupling effect of the channels in neighboring layers directly affects the objective values, and results in a performance gap between (a) and (b). We call this coupling relationship as the \emph{quadratic coupling} between the neighboring layers and formulate the contributions to the objective by quadratic terms of neighboring channel activations. To address this quadratic coupling, we propose a channel selection method based on the QCQP (Quadratic Constrained Quadratic Program) with importance evaluation respecting both the input and the output channels.


\section{Method}
\label{sec:method}
We first propose our discrete QCQP formulation of channel pruning for sequential convolutional neural networks (CNNs). Then, we present two extended versions of our formulation which can handle 1) joint channel and shape pruning of 2D convolution filters and 2) nonsequential connections, respectively.

\subsection{Channel pruning for sequential CNNs}

To capture the importance of weights in $W^{(l)}$, we define the \emph{importance tensor} as $I^{(l)} \in \reals_+^{C_{l\!-\!1}\times C_l \times K_l \times K_l}$. Following the protocol of \citet{han2015learning, guo2016dynamic}, we set $I^{(l)} = \gamma_l \abs{W^{(l)}}$ where $\gamma_l$ is the $\ell_2$ normalizing factor in $l$-th layer or $\|\mathrm{vec}(W^{(l)})\|^{-1}$. Then, we define the binary pruning mask as $A^{(l)} \in \{0,1\}^{C_{l\!-\!1}\times C_l \times K_l \times K_l}$. For channel pruning in sequential CNNs, we define \emph{channel activation} $r^{(l)}\in\{0,1\}^{C_l}$ to indicate which indices of channels remain in the $l$-th layer of the pruned network. Then, the weights in $W^{(l)}_{i,j}$ are active if and only if $r^{(l\!-\!1)}_ir^{(l)}_j=1$, which leads to $A^{(l)}_{i,j} = r^{(l\!-\!1)}_i r^{(l)}_j J_{K_l}$. 
For example, in \Cref{fig:toysub2}, $r^{(l\!-\!1)}=[1,1,1]^\intercal$, $r^{(l)}=[0,1]^\intercal$, and $r^{(l\!+\!1)}=[1,1,0]^\intercal$, therefore, $A^{(l)} = \begin{bmatrix} 0 & 1 \\ 0 & 1 \\ 0 & 1\end{bmatrix} \otimes J_{K_l}$ and $A^{(l\!+\!1)} = \begin{bmatrix} 0 & 0 & 0\\ 1 & 1 & 0\end{bmatrix} \otimes J_{K_{l\!+\!1}}$ \footnote{$\otimes$ denotes the outer product of tensors and $J_n$ is a $n$-by-$n$ matrix of ones.}.

Our goal is to directly maximize the sum of the importance of active weights after the pruning procedure under given resource constraints such as FLOPs, memory usage, network size, or inference time. Concretely, our optimization problem forms as
\begin{align}
    \label{eq:optseq}
    &\maximize_{r^{(0:L)}}~~\sum_{l=1}^L \left\langle I^{(l)}, A^{(l)} \right\rangle \\ 
    &~\mathrm{subject~to} ~~ \sum_{l=0}^L a_l \left\|r^{(l)}\right\|_1 + \sum_{l=1}^L b_l \left\|A^{(l)}\right\|_1 \leq M \nonumber\\
    &\qquad\qquad ~~~~~ A^{(l)} = r^{(l\!-\!1)}{r^{(l)}}^\intercal \otimes J_{K_l} \quad\forall l \in [L].\nonumber
\end{align}
In our formulation, we can exactly compute the actual FLOPS, memory usage, and network size of the pruned network. Furthermore, we can estimate the inference time of the pruned network by modeling it with the number of channels $(=\|r^{(l)}\|_1)$ and the pruning mask sparsity $(=\|A^{(l)}\|_1)$ in each layer, which will be explained in \Cref{subsec:inf}.

The left hand side of the inequality in the first constraint of \Cref{eq:optseq} indicates the actual (or estimated) resource usage. \Cref{tab:rescons} shows the $a_l$ and $b_l$ terms used for computing the usage of each resource. For the $l$-th convolution layer, its network size is equal to the number of parameters in the layer, which is $\|A^{(l)}\|_1$ ($a_l\!\!=\!\!0$, $b_l\!\!=\!\!1$). Memory resource implies the memory used during inference, which is the sum of the memory required for the input feature map, $H_{l\!-\!1} W_{l\!-\!1} \|r^{(l-1)}\|_1 $, and the number of parameters, $\|A^{(l)}\|_1$ ($a_{l\!-\!1}\!\!=\!\!H_{l\!-\!1}W_{l\!-\!1}$, $b_l\!\!=\!\!1$). FLOPs indicates the sum of the number of multiplications for each parameter. Since each parameter requires $H_lW_l$ multiplications, FLOPs equals to $H_l W_l\|A^{(l)}\|_1$ ($a_l\!\!=\!\!0$, $b_l\!\!=\!\!H_lW_l$).

The optimization problem \Cref{eq:optseq} is a discrete nonconvex QCQP of the channel activations $[r^{(0)},\ldots,r^{(L)}]$, where the objective, which is the same with the objective in \Cref{subsec:quad}, respects the quadratic coupling of channel activations $(=r^{(l)})$. Please refer to Supplementary material B for the details on the standard QCQP form of \Cref{eq:optseq}.

\begin{table}[h!]
    \caption{Resource constraints with the corresponding $a_l$ and $b_l$ values. $\beta_l$ and $\delta_l$ are device-specific, and determined via least square regression in \Cref{subsec:inf}.}
\centering
\begin{tabular}{c|c|cc}
\toprule
    Resource constraint (M)& &$a_l$ &$b_l$\\
\midrule
    Network size & exact &$0$     & $1$\\
    Memory       & exact &$H_lW_l$& $1$\\
    FLOPs        & exact &$0$     & $H_lW_l$\\
    \midrule
    Inference time & approx. &$\beta_{l\!+\!1}$     & $\frac{\delta_l}{K_l^2}$\\
\bottomrule
\end{tabular}
    \label{tab:rescons}
\end{table}

\subsection{Inference time constraint}
\label{subsec:inf}

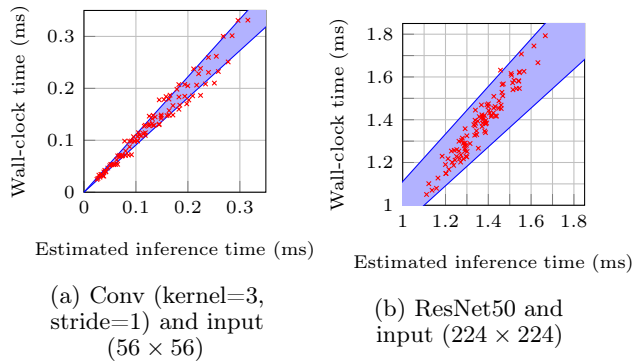
\begin{figure}
\centering
    \captionsetup[subfigure]{justification=centering}
\begin{subfigure}[b!]{0.500\columnwidth}

\begin{tikzpicture}
\begin{axis}[width=4cm, height=4cm, grid=major, scaled ticks = false, ylabel near ticks, tick pos = left, 
tick label style={font=\scriptsize}, 
ytick={0,0.1,0.2,0.3},
xtick={0,0.1,0.2,0.3},
label style={font=\scriptsize}, xlabel={Estimated inference time (ms)}, ylabel={Wall-clock time (ms)}, xmin=0, xmax=0.35, ymin=0, ymax=0.35] 
\addplot+[no markers,name path=A,color=blue,domain=0:1] {1/0.9*\x};
\addplot+[no markers,name path=B,color=blue,domain=0:1] {1/1.1*\x};
\addplot+[red, only marks, mark options={fill=red, scale=0.6, mark=x, solid}] table [x=est, y=inf, col sep=comma]{csv/inf_exps/sol1.csv};
\addplot[no markers,blue!30] fill between[of=A and B];

\end{axis}
\end{tikzpicture}

\centering
    \caption{Conv (kernel=3, stride=1) and input ($56\times56$)}
\label{fig:infexpconv}
\end{subfigure}
\begin{subfigure}[b!]{0.48\columnwidth}

\begin{tikzpicture}
\begin{axis}[width=4cm, height=4cm, grid=major, scaled ticks = false, ylabel near ticks, tick pos = left, 
tick label style={font=\scriptsize}, 
ytick={1,1.1,1.2,1.3,1.4,1.5,1.6,1.7,1.8},
yticklabels={1,,1.2,,1.4,,1.6,,1.8},
xtick={1,1.1,1.2,1.3,1.4,1.5,1.6,1.7,1.8},
xticklabels={1,,1.2,,1.4,,1.6,,1.8},
label style={font=\scriptsize}, xlabel={Estimated inference time (ms)}, ylabel={Wall-clock time (ms)}, xmin=1, xmax=1.85, ymin=1, ymax=1.85] 
\addplot+[no markers,name path=A,color=blue,domain=0:2] {1/0.9*\x};
\addplot+[no markers,name path=B,color=blue,domain=0:2] {1/1.1*\x};
\addplot+[red, only marks, mark options={fill=red, scale=0.6, mark=x, solid}] table [x=estimate, y=real, col sep=comma]{csv/infer_scatter50_0.csv};
\addplot[no markers,blue!30] fill between[of=A and B];

\end{axis}
\end{tikzpicture}
\centering
\caption{
    ResNet50 and \\input ($224\times224$)
}
\label{fig:infexpnet}
\end{subfigure}

\caption{
    Estimated inference time vs.\ actual wall-clock time of a single convolution operation (a) and a ResNet-50 (b) while varying the number of channels. 
    The blue area indicates the area where the error of the estimated value is under 10\% with respect to the actual wall-clock time. The wall-clock time is measured on a machine with Intel Xeon E5-2650 CPU and Titan XP GPU. Additional experiment results are provided in Supplementary material D.
}
\label{fig:infexp}
\end{figure}
Unlike other resource constraints, inference time on the edge-device can only be measured by running each pruned network on the real device. However, deploying every candidate networks on the device during pruning can be prohibitive. Therefore, we propose an approximate method that accurately predicts the inference time of the pruned networks efficiently.

We utilize the fact that the computation cost of a pruned network is highly dependent to the compution cost of convolution operations with the pruned channels. We first build a quadratic model with respect to the number of unpruned channels, $\|r^{(l)}\|_1$, to estimate the inference time of each convolution operation. Then, we estimate the inference time of the pruned network by integrating these estimate values.

Concretely, we model the inference time of the $l$-th layer convolution operation with respect to the number of its input channels ($=\|r^{(l\!-\!1)}\|_1$) and the output channels ($=\|r^{(l)}\|_1$). 
Using these variables, we aim to model three major factors of the convolution operation that affect the inference time: 1) FLOPs (computation cost), 2) MAC (memory access cost), and 3) bias overhead. 
The resulting estimation model for the $l$-th convolutional operation is $\alpha_l+\beta_l\|r^{(l\!-\!1)}\|_1+\delta_l\|r^{(l\!-\!1)}\|_1\|r^{(l)}\|_1$, where each terms represent the contribution of bias overhead, MAC, and FLOPs to the inference time. More discussions for choosing this model and analysis of the coefficients $\alpha_l$, $\beta_l$, and $\delta_l$ are provided in Supplementary material D. 
Note that $\alpha_l$, $\beta_l$, and $\delta_l$ are dependent on the edge-device. Therefore, we find the best $\alpha_l$, $\beta_l$, and $\delta_l$ values via least square regression on a few samples of $(\|r^{(l\!-\!1)}\|_1, \|r^{(l)}\|_1, \text{WallClock}^{(l)}(r^{(l\!-\!1)}, r^{(l)}))$, where $\text{WallClock}^{(l)}$ denotes the wall-clock time of the $l$-th convolution operation with $\|r^{(l\!-\!1)}\|_1$ input channels and $\|r^{(l)}\|_1$ output channels. Finally, we plug the learned quadratic model into the left hand side of the inequality in \Cref{eq:optseq} \footnote{Concretely, we set $a_l$ and $b_l$ of  \Cref{eq:optseq} to $\beta_{l\!+\!1}$ and $\delta_l/K_l^2$. $K_l^2$ in the denominator comes from $A^{(l)} = K_l^2\|r^{(l\!-\!1)}\|_1\|r^{(l)}\|_1$.}. In this paper, we measure the actual wall-clock time using Nimble framework \citep{nimble} on CUBLAS backend.

\Cref{fig:infexp} shows the prediction performance of our quadratic model. The red points represent the estimated inference time versus the actual inference time of a single convolution operation (a) and a ResNet-50 network (b) while varying the number of channels. Our model successfully estimates the actual inference time of single convolution operations with $8\%$ error rate\footnote{We evaluate the mean percent error by $\frac{1}{n}\sum_{i=1}^n\frac{|E_i-A_i|}{A_i}$ where $E_i$ and $A_i$ are the estimated and the actual value, respectively.} on average. Also, we estimate the inference time of the whole network by summing all of the estimated inference time of its convolution operations, and achieve $3\%$ error rate on average. Note that as we gather the inference time of the convolution operations, the variance of the inference time decreases by the law of large number and the error rate simply decreases.

\subsection{Joint channel and spatial pruning}

For further efficiency, we increase the pruning granularity and jointly perform spatial pruning to 2-D convolution filters. Concretely, we prune by each weight tensor across the input channel direction additionally to perform channel and spatial pruning processes simultaneously.

\input{fig/pointwise_3D_bold2.tex}

First, we define the \emph{shape column} $W^{(l)}_{\cdot, j, a, b}$ by the vector of weights at spatial position $(a,b)$ of a 2-D convolution filter along the $j$-th output channel dimension. Then, we define \emph{shape column activation} $q^{(l)} \in \{0,1\}^{C_l\times K_l\times K_l}$ to indicate which shape columns in the $l$-th convolution layer remain in the pruned network. \Cref{fig:point3d} shows the illustration of each variables.

Note that this definition induces constraints on the channel activation variables. In detail, the $j$-th output channel activation in $l$-th layer is set if and only if at least one shape column activation in the $j$-th output channel is set. Concretely, the new formulation should include the constraints $r^{(l)}_j \leq \sum_{a,b} q^{(l)}_{j,a,b}$ and $q^{(l)}_{j,a,b} \leq r_j^{(l)}~ \forall a,b$. Our optimization problem with these shape column activation variables are introduced in Supplementary material A.

We can also formulate this optimization problem as a discrete nonconvex QCQP. The details on the standard QCQP form are provided in Supplementary material B. Furthermore, we show that the constraints in our optimization problems provably eliminate any unpruned inactive weights and accurately model the resource usage as well as the objective of the pruned network. The proof of this statement is given in Supplementary material C.

\subsection{Handling nonsequential connections}
\label{subsec:nonseq}

\input{fig/nonseq.tex}

Nonsequential connections, such as skip additions \citep{resnet, howard2017mobilenets, efficientnet} or skip concatenations \citep{densenet}, are an essential part of modern neural networks. However, previous pruning works focus on dealing with only sequential connections and often resort to simple heuristics \citep{liu2017learning, sfp,fpgm,molchanov2016pruning, molchanov2019importance}. In this subsection, we show that our method can be naturally generalized to handle nonsequential connections, providing a much larger optimization search space compared to the previous works. As an example, we compare our method with the heuristic adopted by GBN \citep{you2019gate} in \Cref{fig:nonseq}.

Before going into the details, we first define some new notations in \Cref{tab:notation}, as an output feature map does not directly correspond to the subsequent layer's input feature map when nonsequential connections exist. First, we denote the input feature map and output feature map of the $t$-th convolution layer as $V^{(t\!-\!1)} \in \mathbb{R}^{C_{t\!-\!1}\times H_{t\!-\!1} \times W_{t\!-\!1}}$ and $U^{(t)} \in \mathbb{R}^{C_{t}\times H_{t} \times W_{t}}$, respectively. Then, we define input channel activation $v^{(t\!-\!1)} \in \{0,1\}^{C_{t\!-\!1}}$ and output channel activation $u^{(t)}\in \{0,1\}^{C_{t}}$ to indicate the remaining output channels of the  $V^{(t\!-\!1)}$ and $U^{(t)}$ after pruning.

\begin{table}[h!]
	\caption{Notation change of feature map and channel activation from sequential CNNs to nonsequential CNNs.}
	\centering
	\begin{tabular}{cccc}
		\toprule
		CNN type&Sequential& \multicolumn{2}{c}{Nonsequential}\\
		\cmidrule(r){3-4}
		&           &input &output \\
		\midrule
		Feature map& $X^{(l)}$ & $V^{(l)}$ & $U^{(l)}$\\
		Channel activation & $r^{(l)}$ & $v^{(l)}$ & $u^{(l)}$\\
		\bottomrule
	\end{tabular}
	\label{tab:notation}
\end{table}

Assume layer $s$ and $t$ are nonsequentially connected via skip addition, as illustrated in \Cref{fig:nonseq}. Since $v^{(s)}$ and $u^{(t)}$ both affect $v^{(t)}$, we need a protocol to resolve $v^{(t)}$ when $v^{(s)}$ and $u^{(t)}$ are set differently. To handle this problem, GBN simply imposes the constraint $v^{(s)}=u^{(t)}=v^{(t)}$ to avoid any conflicts between the output channel activations. On the other hand, our method allows $v^{(s)}$ and $v^{(t)}$ to differ, and resolve the conflicts algorithmically according to the status of each activation masks. As a result, our method is able to adopt a more flexible constraint: $u^{(t)} \preceq v^{(t)} \preceq u^{(t)}+v^{(s)}$. More details of this procedure and the resulting optimization form are provided in Supplementary material A. When $v^{(s)}$, $u^{(t)}$, and $v^{(t)}$ are of dimension $n$, the number of possible $(v^{(s)}, u^{(t)}, v^{(t)})$ combinations without any constraints is $8^n$. The constraint of GBN reduces this number to $2^n$, while our constraint reduces the number to only $5^n$. This shows our method provides a much larger optimization search space compared to the previous method.

We generalize \Cref{eq:optseq} to other types of nonsequential connections such as skip concatenations or skip additions with different dimensions in Supplementary material A. Note that the constraints in the generalized \Cref{eq:optseq} also eliminate any unpruned inactive weights in nonsequential networks with skip additions. The proof of this statement is also given in Supplementary material C.

\subsection{Objective optimization}
\Cref{eq:optseq} and generalized optimization problems of \Cref{eq:optseq} fall into the category of binary Mixed Integer Quadratic Constraint Quadratic Programming (MIQCQP). We solve these discrete QCQP problems with the CPLEX library \citep{Cplex}, which provides MIQCQP solvers based on the branch and cut technique. However, the branch and cut algorithm can lead to exponential search time \citep{bb} on large problems. Therefore, we provide a practical alternative utilizing a block coordinate descent style optimization, described in Supplementary material E. 

\section{Related works}


\paragraph{Importance of channels} \label{par:imp} Most of the channel pruning methods prune away the least important channels with a simple greedy approach, and the evaluation method for the importance of channels has been the main research problem \citep{molchanov2016pruning, molchanov2019importance, liu2019}. Channel pruning is divided into two major branches according to the method of evaluating the importance of channels: the \emph{trainable-importance} method, which constantly evaluates the importance of channels while training the whole network from scratch, and the \emph{fixed-importance} method, which directly evaluates the importance of channels on the pretrained network. Trainable-importance channel pruning methods include regularizer-based methods with group sparsity regularizers \citep{wen2016learning, alvarez2016learning, yang2019deephoyer, liu2017learning, louizos2017learning, gordon2018morphnet} and data-driven channel pruning methods \citep{kang2020operationaware, you2019gate}. Fixed-importance channel pruning methods first prune away most of the weights and then finetune the significantly smaller pruned network \citep{molchanov2016pruning, molchanov2019importance, apoz, sfp, pfec, fpgm, luo2017thinet,ccp}. As a result, fixed-importance methods are much more efficient than the trainable-importance methods in terms of computational cost and memory usage as trainable-importance methods require training the whole unpruned network. Our framework is on the line of fixed-importance channel pruning works. For example, when pruning a ResNet for CIFAR-10 dataset, \citet{liu2017learning} trains the entire network with a sparsity regularizer for 160 epochs, prunes the trained network, and then finetunes it for another 160 epochs. Meanwhile, our method only requires pruning the pretrained network and finetuning the the pruned network for 200 epochs.

\paragraph{Quadratic coupling} CCP \citep{ccp} formulates a QP (quadratic formulation) to consider the quadratic coupling between channels in the same layer under predefined layer-wise constraints on the maximum number of channels. On the other hand, our formulation considers the quadratic coupling between channels in the neighboring layers under the target resource constraints.


\paragraph{Spatial pruning} Spatial pruning methods aim to prune convolution filters along the channel dimension for inference efficiency. Spatial pruning methods manually define the spatial patterns of filters \citep{lebedev2016fast, anwar2017structured} or optimize spatial patterns of filters with group sparse regularizers \citep{wen2016learning, lebedev2016fast}. Among these works, \citet{lebedev2016fast} empirically demonstrates that enforcing sparse spatial patterns in 2-D filters along the input channel leads to great speed-up during inference time using group sparse convolution operations \citep{chellapilla2006high}. Our proposed method enforces the spatial patterns in 2-D filters as in \citet{lebedev2016fast} for speed-up in inference.

\paragraph{Inference time constraint} 

Even though inference time is a metric that many machine learning practitioners are interested in, previous pruning methods resort to a well-known but inaccurate proxy, FLOPs, due to the difficulty of measuring and modeling the inference time. NetAdapt and AMC \citep{yang2018netadapt, he2019amc} directly measure the inference time of proposed networks by deploying each of them on the edge device. However, this approach incurs a high cost since we have to place, run, and remove every candidate network. On the other hand, we propose a quadratic model to estimate the inference time of the whole pruned network without direct deployment.

\section{Experiments}
We compare the classification accuracy of the pruned network against several pruning baselines on CIFAR-10 and ImageNet datasets using DenseNet-40 \citep{densenet}, VGG-16 \citep{simonyan2015deep}, EfficientNet \citep{efficientnet}, and various versions of ResNet \citep{resnet}. Note that most pruning baselines apply an iterative pruning procedure, which repeatedly alternates between network pruning and finetuning until the target resource constraints are satisfied \citep{han2015learning, sfp, liu2017learning, yang2018netadapt}. In contrast, our methods explicitly combine the target resource constraint to the optimization framework and only need one round of pruning and finetuning.

\subsection{Experimental details} 
\label{subsec:expset}
We follow the `smaller-norm-less-important' criterion \citep{ye2018rethinking, liu2017learning}, which evaluates the importance of weights with the absolute value of the weight \citep{han2015learning, guo2016dynamic}.
We assume FLOPs reduction are linearly proportional to the sparsity in shape column activations, as empirically shown in \citet{lebedev2016fast}. In the experiment tables, FLOPs of the pruned network are computed according to the resource specifications in \Cref{eq:optseq}. The `IC' column indicates whether each method is a trainable-importance method (T) or a fixed-importance method (F).  `ours-c' and `ours-cs' each refers to our method with only channel pruning and with both the channel and spatial pruning, respectively. The experiment results on network size constraints are provided in Supplementary material F.

\begin{table}[!ht]
    \caption{Pruned accuracy and accuracy drop from the baseline network at given FLOPs on various network architectures (ResNet-20,32,56 and DenseNet-40) at CIFAR-10. Asterisk (*) indicates that the method does not use finetuning.} 
	\centering
	\begin{adjustbox}{max width=\columnwidth}
        \setlength\tabcolsep{4pt}
        \begin{tabular}{lccccc}
            \toprule
            Method&IC&Baseline acc&Pruned acc$\uparrow$&Acc drop$\downarrow$&FLOPs(\%)$\downarrow$\\
            \midrule
            \multicolumn{3}{l}{Network: ResNet-20}\\
            \midrule
            SFP \citep{sfp}  &F&92.20 (0.18)  &90.83 (0.31)           &1.37           &\textbf{57.8}\\ 
            FPGM \citep{fpgm}&F&92.21 (0.18)  &91.72 (0.20)           &0.49           &\textbf{57.8}\\             
            ours-c &F&92.21 (0.18)  &91.74 (0.20)           &0.47           &58.3\\
            ours-cs &F&92.21 (0.18)  &\textbf{92.26} (0.10)  &\textbf{-0.05} &\textbf{57.8}\\
            \midrule
            \midrule
            \multicolumn{3}{l}{Network: ResNet-32}\\
            \midrule
            SFP \citep{sfp}         &F&92.63 (0.70)  &92.08 (0.08)          &0.55              &58.5\\ 
            FPGM \citep{fpgm}       &F&92.88 (0.86)  &92.51 (0.90)          &0.37               &58.5\\
            ours-c         &F&92.88 (0.86)	 &92.52 (0.46)	        &0.36	           &\textbf{57.2}\\
            ours-cs       &F&92.88 (0.86)	 &\textbf{92.80} (0.61)	&\textbf{0.08}     &57.9         \\ 
            \midrule
            \midrule
            \multicolumn{3}{l}{Network: ResNet-56}\\
            \midrule
            SFP \citep{sfp}     &F&93.59 (0.58)  &92.26 (0.31)           &1.33              &47.5\\ 
            FPGM \citep{fpgm}   &F&93.59 (0.58)  &93.49 (0.13)           &0.10              &47.5\\
            CCP \citep{ccp}	&F&93.50 &93.42 &0.08&\textbf{47.4}\\
            SCP \citep{kang2020operationaware}	&T*&93.69		&93.23			&0.46		&48.5\\			            
            ours-c  &F&93.59 (0.58)  &93.36 (0.68)           &0.23              &\textbf{47.4}\\
            ours-cs &F&93.59 (0.58)  &\textbf{93.59} (0.36)  &\textbf{0.00}     &\textbf{47.4}\\
            \midrule
            \midrule
            \multicolumn{3}{l}{Network: DenseNet-40}\\
            \midrule
            SCP \citep{kang2020operationaware}&T*&94.39&93.77&\textbf{0.62}&\textbf{29.2}\\
            ours-c&F&95.01&93.80&1.21&\textbf{29.2}\\
            ours-cs&F&95.01&\textbf{94.25}&0.76&\textbf{29.2}\\  
            \midrule
            slimming \citep{liu2017learning}&T&93.89&94.35&\textbf{-0.46}&\textbf{45.0}\\
            ours-c&F&95.01&94.38&0.63&\textbf{45.0}\\
            ours-cs&F&95.01&\textbf{94.85}&0.16&\textbf{45.0}\\      
            \midrule
            slimming \citep{liu2017learning}&T&93.89&94.81&\textbf{-0.92}&71.6\\
            ours-c&F&95.01&94.82&0.19&\textbf{71.0}\\
            ours-cs&F&95.01&\textbf{95.02}&-0.01&\textbf{71.0}\\
            \bottomrule
		\end{tabular}
	\end{adjustbox}
	\label{tab:cifar10}
\end{table}

\subsection{CIFAR-10}
CIFAR-10 dataset has $10$ different classes with $5k$ training images and $1k$ test images per each class \cite{cifar}. In CIFAR-10 experiments, we evaluate our methods on four network architectures: ResNet-20, 32, 56, and DenseNet-40. Implementation details of our experiments are listed in Supplementary material F. We show the experiment results of pruning under FLOPs constraints in \Cref{tab:cifar10}.

We find that `ours-c' shows comparable results against FPGM, which is the previous state of the art method, on ResNet-20, 32, and 56. Moreover, `ours-cs' significantly outperforms both `ours-c' and FPGM on the same architectures, showing a state of the art performance. Also, `ours-c' shows comparable results against slimming \citep{liu2017learning} and SCP \citep{kang2020operationaware}, which are trainable-importance methods, while `ours-cs' outperforms the baselines by a large margin on DenseNet-40. These results show simultaneous channel and spatial pruning produces more computationally efficient networks with better performance compared to other channel pruning methods on CIFAR-10.
\begin{table}[!hbt]
    \caption{Top1,5 pruned accuracy and accuracy drop from the baseline network at given FLOPs on various network architectures (ResNet-18, 50, VGG-16, and EfficientNet-B0) at ImageNet. \dag\ denotes the methods which report their results to the first decimal place.}
	\centering
	\begin{adjustbox}{max width=\columnwidth}
        \setlength\tabcolsep{4pt}
		\begin{tabular}{lcccc}
			\toprule 
            Method&IC&Top1 Pruned Acc$\uparrow$& Top1 Acc drop$\downarrow$&FLOPs(\%)$\downarrow$\\ 
            \midrule 
            \multicolumn{3}{l}{Network: ResNet-18}\\
            \midrule
            SFP \citep{sfp}    &F  &67.10            &3.18           &\textbf{58.2} \\
            FPGM \citep{fpgm}  &F  &68.41            &1.87           &\textbf{58.2} \\    
            ours-c             &F  &67.48	          &2.28           &60.9 \\
            ours-cs            &F  &\textbf{69.59}  &\textbf{0.17}  &\textbf{58.2} \\ 
       	    \midrule
            \midrule
            \multicolumn{3}{l}{Network: ResNet-50}\\
            \midrule
            SFP \citep{sfp} &F&74.61           &1.54          &58.3 \\
            FPGM \citep{fpgm} &F&75.50           &0.65          &\textbf{57.8} \\
            ours-c          &F &75.78           &0.37          &\textbf{57.8} \\ 
            ours-cs         &F &\textbf{75.93}	 &\textbf{0.22} &\textbf{57.8} \\ 
            \midrule
            GBN \citep{you2019gate}	& T&\textbf{76.19}		&\textbf{-0.31}	&59.5\\
            ours-c                  & F&75.89              &0.26           &61.5 \\ 
            ours-cs                 & F&76.00              &0.15           &\textbf{59.0} \\ 
            \midrule
            \midrule
            \multicolumn{3}{l}{Network: EfficientNet-B0}\\
            \midrule
            uniform MP& F&75.06&2.57&\textbf{76.4}\\	
            ours-c&F&\textbf{75.71}&\textbf{1.92}&\textbf{76.4}\\         
            \midrule
            uniform MP&F&69.08&8.55&53.8\\	
            ours-c&F&\textbf{73.27}&\textbf{4.36}&\textbf{53.7}\\
            \midrule
            \midrule
            Method&IC&Top5 Pruned Acc$\uparrow$& Top5 Acc drop$\downarrow$&FLOPs(\%)$\downarrow$\\ 
            \midrule
            \multicolumn{3}{l}{Network: VGG-16}\\
            \midrule
            \citet{molchanov2016pruning}\dag&F&84.5&5.9&\textbf{51.7}\\	
            ours-c                      &F&87.20&3.18&\textbf{51.7}\\
            ours-cs                     &F&\textbf{87.36}&\textbf{3.02}&\textbf{51.7}\\
        \bottomrule
		\end{tabular}
	\end{adjustbox} 
    \label{tab:imgnet}
\end{table}

\subsection{ImageNet}
ILSVRC-2012 \citep{imagenet} is a large-scale dataset with $1000$ classes that comes with $1.28M$ training images and $50k$ validation images. 
We conduct our methods under the fixed FLOPs constraint on ResNet-18, 50, EfficientNet-B0, and VGG-16. For more implementation details of the ImageNet experiments, refer to Supplementary material F. We show ImageNet experiment results in \Cref{tab:imgnet}. In ResNet-50, `ours-c' and `ours-cs' achieve results comparable to GBN, a trainable-importance channel pruning method which is the previous state of the art, even though our method is a fixed-importance channel pruning method. In particular, top1 pruned accuracy in `ours-cs' exceeds SFP by $1.32\%$ using a similar number of FLOPs. Both `ours-cs' and `ours-c' clearly outperform FPGM in ResNet-50. In EfficientNet-B0, we compare `ours-c' with `uniform MP', where `uniform MP' denotes a magnitude-based channel pruning method which greedily prunes filters with small weight norms uniformly among layers. `ours-c' again outperforms `uniform MP' in various FLOPs constraints. Also, `ours-c' and `ours-cs' show significantly better performance compared to \citet{molchanov2016pruning} on VGG-16. More experiments on FLOPs constraints with MobileNetV2 and image segmentation tasks are provided in Supplementary F.
\begin{table}[h!]
    \caption{Top1 pruned accuracy and accuracy drop from the baseline network at given inference time on ResNet-50 architecture at ImageNet.}
\centering
\begin{adjustbox}{max width=1.0\columnwidth}
\begin{tabular}{lcccc}
\toprule
Method&IC& Top1 Pruned Acc$\uparrow$ & Top1 Acc drop$\downarrow$ & Inference time (ms)$\downarrow$ \\
\midrule
FPGM&F&75.50&0.65&1.68 (1.51$\times$)\\
ours-c&F&\textbf{75.83}&\textbf{0.32}&\textbf{1.66} (\textbf{1.52$\times$})\\
\bottomrule
\end{tabular}
\end{adjustbox}
    \label{tab:infer}
\end{table}

\Cref{tab:infer} shows the pruning result under inference time constraints. We observe that `ours-c' outperforms FPGM on ResNet-50 in both top1 pruned accuracy and the inference time, reducing the accuracy drop to half compared to FPGM. Estimating the inference time of the pruned network using the quadratic model, our method successfully finds a network with faster inference than the baseline without direct deployment.

\section{Conclusion}

We propose an optimal channel selection method with a discrete QCQP based optimization framework. Greedy channel selection methods ignore the inherent quadratic coupling between channels in the neighboring layers and fail to eliminate inactive weights during pruning. To this end, our selection method models the quadratic coupling explicitly and prevents any inactive weights during the pruning procedure. Our selection method allows exact modeling of the user-specified resource constraints in terms of FLOPS, memory usage, and network size, which enables the direct optimization of the true objective on the pruned network. In addition, we propose a new quadratic model that accurately estimates the inference time of a pruned network without direct deployment, which allows us to adopt inference time as a resource constraint option. We also extend our method to also select individual 2D convolution filters simultaneously and handle nonsequential operations in modern neural networks more flexibly.  Extensive experiments show our proposed method significantly outperforms other fixed-importance channel pruning methods, finding smaller and faster networks with the least drop in accuracy.

\newpage

\section*{Acknowledgement}
This research was supported in part by Samsung Advanced Institute of Technology, Samsung Electronics Co., Ltd, Institute of Information \& Communications Technology Planning \& Evaluation (IITP) grant funded by the Korea government (MSIT) (No. 2020-0-00882, (SW STAR LAB) Development of deployable learning intelligence via self-sustainable and trustworthy machine learning), and Basic Science Research Program through the National Research Foundation of Korea (NRF) (2020R1A2B5B03095585). Yeonwoo Jeong was supported by NRF(National Research Foundation of Korea) Grant funded by the Korean Government(NRF-2019-Global Ph.D. Fellowship Program). Hyun Oh Song is the corresponding author.

\bibliography{aistats}
\bibliographystyle{apalike}

\newpage 

\onecolumn
\aistatstitle{Supplementary Material}

\appendix
\section{QCQP formulation on Nonsequential connections}

We first formulate the optimization problem for channel and spatial pruning in Section 3.3. Then, we generalize the optimization problem to cover nonsequential connections. 

To recap, the original formulation of our optimization problem is as follows:
\begin{align}
    \label{eq:optseq}
    &\maximize_{r^{(0:L)}}~~\sum_{l=1}^L \left\langle I^{(l)}, A^{(l)} \right\rangle \\ 
    &~\mathrm{subject~to} ~~ \sum_{l=0}^L a_l \left\|r^{(l)}\right\|_1 + \sum_{l=1}^L b_l \left\|A^{(l)}\right\|_1 \leq M \nonumber\\
    &\qquad\qquad ~~~~~ A^{(l)} = r^{(l\!-\!1)}{r^{(l)}}^\intercal \otimes J_{K_l} \quad\forall l \in [L].\nonumber
\end{align}

Concretely, we handle two prevalent types of nonsequential connections: skip addition \citep{resnet, howard2017mobilenets, efficientnet} and skip concatenation \citep{densenet}.

\vspace{2em}
\textbf{ A.1 Joint channel and spatial pruning}
\vspace{2em}

We first recap the definition of shape column activation. The shape column activations $q^{(l)} \in \{0,1\}^{C_l\times K_l\times K_l}$ indicate which shape columns in the $l$-th convolution layer remain in the pruned network. 
Then, the new formulation include the constraints $r^{(l)}_j \leq \sum_{a,b} q^{(l)}_{j,a,b}$ and $q^{(l)}_{j,a,b} \leq r_j^{(l)}~ \forall a,b$. 

We aim to maximize the sum of the importance of active weights after pruning under the given resource constraints. Then, our optimization problem for joint channel and spatial pruning becomes 

\begin{align}
    \label{eq:jointoptseq}
    &\maximize_{r^{(0:L)}, q^{(1:L)}} ~~\sum_{l=1}^L \left\langle I^{(l)}, A^{(l)} \right\rangle \\
    &~\mathrm{subject~to~~} \sum_{l=0}^{L} a_l \left\|r^{(l)}\right\|_1 + \sum_{l=1}^L b_l \left\|A^{(l)}\right\|_1 \leq M \nonumber\\
    &\qquad \qquad \quad r^{(l)}_j \leq \sum_{a,b} q^{(l)}_{j,a,b}\quad q^{(l)}_{j,a,b} \leq r_j^{(l)} \quad \forall l,j,a,b\nonumber\\
    &\qquad \qquad \quad A^{(l)} = r^{(l\!-\!1)} \otimes q^{(l)} \quad \forall l \nonumber\\
    &\qquad \qquad \quad r^{(l)} \in \{0,1\}^{C_l} \quad q^{(l)} \in \{0,1\}^{C_l\times K_l\times K_l} \quad \forall l \in [L].\nonumber
\end{align}

\newpage
\textbf{ A.2 Skip addition}
\vspace{2em}


We first introduce two notations, \emph{output channel activation} $u^{(t)} \in \{0,1\}^{C_t}$ and \emph{input channel activation} $v^{(t)} \in \{0,1\}^{C_t}$, along with the corresponding new constraints, in \Cref{tab:notation2}. As described in Section 3.4, the definition of $u^{(t)}$ induces new constraints between the shape column activations $\left(=q^{(t)} \right)$ and output channel activations $\left(=u^{(t)}\right)$. Concretely, the constraints are given as $u^{(t)}_j \leq \sum_{a,b} q^{(t)}_{j,a,b}$ and $q^{(t)}_{j,a,b} \leq u_j^{(t)} ~\forall a,b$. 


We now discuss the constraints between the input and output channel activation variables under four possible scenarios depending on the architectural implementations of skip additions. Here, we denote the set of layer index pairs $\{(s,t)\}$ which have skip additions as $\mathcal{P}$. Concretely, $(s, t)\in \mathcal{P}$ if and only if the input feature map of the $s\!+\!1$-th convolution layer is added to the output feature map of the $t$-th layer, forming the input feature map of the $t\!+\!1$-th layer. Also, let $T=\{t\mid (s,t)\in \mathcal{P}\}$. For a layer $t$, we formulate the channel activation constraints for each possible connection scenarios separately:

$(i)$ If there is no skip addition incoming to the $t$-th layer ($t\notin T$), then we force $u^{(t)}=v^{(t)}$. 
\vspace{1em}

$(ii)$ For a skip addition pair $(s,t)\in \mathcal{P}$ with matching channel dimensions ($C_s\!=\!C_t$), the input feature map of the $s\!+\!1$-th convolution layer is directly added to the output feature map from the $t$-th layer as illustrated in \Cref{fig:ef2}. In this case, we can formulate the constraints as $u^{(t)} \preceq v^{(t)} \preceq u^{(t)} + v^{(s)}$.
\vspace{1em}

$(iii)$ For a skip addition pair $(s,t)\in \mathcal{P}$ with mismatching channel dimensions ($C_s\!<\!C_t$), the skip addition can utilize zero padding $(iii-a)$ or $1\times1$ convolutions ($iii-b$) to resolve the mismatch \citep{resnet}. We define the \emph{augmented feature map} $\tilde{V}^{(s)}$ after the zero padding or $1\!\times \!$ convolution and corresponding \emph{augmented channel activation} $\tilde{v}^{(s)}\in \{0,1\}^{C_t}$. Then, we formulate the constraints for both cases as below. Not that similar with the constraint in $(ii)$, the constraints for both cases are formulated as $u^{(t)} \preceq v^{(t)} \preceq u^{(t)} + \tilde{v}^{(s)}$.
\vspace{1em}

\input{supp_fig/nonseq.tex}

$~~~~(iii-a)$
        A ($C_{t}\!-\!C_{s}$)-dimensional zero-valued feature map is padded to the end of the $s\!+\!1$-th convolution layer's input feature map. We define $\tilde{v}^{(s)} = \left[v^{(s)}, 0_{C_t-C_s}\right]$, as illustrated in \Cref{fig:ef3_type1}. Therefore, for all $j\leq C_s$, $u^{(t)}_j \leq v^{(t)}_j \leq u^{(t)}_j + v^{(s)}_j$ and for all $j> C_s$, $u^{(t)}_j \leq v^{(t)}_j \leq u^{(t)}_j + 0$.

$~~~~(iii-b)$
      $1\times 1$ convolution is applied to the $s\!+\!1$-th layer's input feature map to match the larger channel dimension $\left(=C_{t}\right)$. Since the number of FLOPs and weights in a $1\times 1$ convolution is negligible compared to the total number of FLOPs and weights, we assume all of the output channels of a $1\times 1$ convolution are activated and define $\tilde{v}^{(s)} = 1_{C_t}$ as illustrated in \Cref{fig:ef3_type2}. Therefore, $u^{(t)} \preceq v^{(t)} \preceq u^{(t)} + 1_{C_t}$.

\begin{table}[h]
    \caption{Notation and constraints change of feature map and channel activation from sequential CNNs to nonsequential CNNs with skip addition.}
	\centering
    \begin{adjustbox}{max width=\columnwidth}
	\begin{tabular}{lccc}
		\toprule
		CNN type&Sequential& \multicolumn{2}{c}{Nonsequential}\\
		\cmidrule(r){3-4}
		&           &input &output \\
		\midrule
        Notation\\
        \midrule
		Feature map& $X^{(l)}$ & $V^{(l)}$ & $U^{(l)}$\\
		Channel activation & $r^{(l)}$ & $v^{(l)}$ & $u^{(l)}$\\
        \midrule
        Constraint\\
        \midrule
        Binary pruning mask & $A^{(l)}=r^{(l\!-\!1)}\otimes q^{(l)}$ & \multicolumn{2}{c}{$A^{(l)}=v^{(l\!-\!1)} \otimes q^{(l)}$}\\

        Shape column activation & $r^{(l)}_j \leq \sum_{a,b} q^{(l)}_{j,a,b}$ & \multicolumn{2}{c}{$u^{(l)}_j \leq \sum_{a,b} q^{(l)}_{j,a,b}$}\\
                                & $q^{(l)}_{j,a,b} \leq r_j^{(l)}$ & \multicolumn{2}{c}{ $q^{(l)}_{j,a,b} \leq u_j^{(l)}$ }\\
		\bottomrule
	\end{tabular}
    \end{adjustbox}
	\label{tab:notation2}
\end{table}

\newpage

We now summarize the constraints for the four cases discussed above to \Cref{eq:efconstraintshort}.
\begin{align}
    &(i)~~  v^{(t)} = u^{(t)} \quad \forall t \notin T \nonumber \\
    &(ii)~~  u^{(t)} \preceq v^{(t)} \preceq u^{(t)} + v^{(s)} \quad \forall (s,t) \in \mathcal{P} \text{ and } C_t=C_s\nonumber\\
    &(iii)~~ \forall (s,t) \in \mathcal{P} \text{ and } C_t>C_s,\nonumber\\ 
    &\qquad(iii-a)~~ u^{(t)}_j \leq v^{(t)}_j \leq u^{(t)}_j + v^{(s)}_j \quad \forall j \leq C_s\nonumber\\
    &\qquad \qquad \qquad \text{ and } v^{(t)}_j = u^{(t)}_j \quad \forall j > C_{s}\nonumber\\
    &\qquad(iii-b)~~ u^{(t)} \preceq v^{(t)}.
    \label{eq:efconstraintshort}
\end{align}
In Supplementary material C, we prove that the constraints of \Cref{eq:efconstraintshort} prevent inactive weights from remaining in the pruned network with skip additions.

We now formulate the network channel and spatial pruning optimization problem that handles nonsequential connections using the input channel, output channel, and shape column activation variables:

\footnotesize
\begin{align}
    \label{eq:efjointopt}
    &\maximize_{u^{(0:L)},v^{(0:L)},q^{(1:L)}} ~~\sum_{t=1}^L \left \langle I^{(t)}, A^{(t)}\right\rangle \\
    &\quad~~\mathrm{subject~to} \nonumber\\
    &\qquad~~ \sum_{t=0}^{L} a_t \left\|u^{(t)}\right\|_1 + \sum_{t\in T} a_t \left\|v^{(t)}\right\|_1 + \sum_{t=1}^L b_t \left \|A^{(t)}\right\|_1 \leq M \qquad~~ (i)~~  v^{(t)} = u^{(t)} \quad \forall t \notin T \nonumber\\
    &\qquad~~ u^{(t)}_j \leq \sum_{a,b} q^{(t)}_{j,a,b}  ~~~\text{and}~~~ q^{(t)}_{j,a,b} \leq u_j^{(t)} \quad \forall t,j,a,b \qquad\qquad~~~~ (ii)~~~~~  u^{(t)} \preceq v^{(t)} \preceq u^{(t)} + v^{(s)} \quad \forall (s,t) \in \mathcal{P} \text{ and } C_t=C_s\nonumber\\
    &\qquad~~ A^{(t)} = v^{(t\!-\!1)}\otimes q^{(t)} \quad \forall t  \qquad \qquad\qquad\qquad\qquad\qquad\qquad~~~~ (iii)~~ \forall (s,t) \in \mathcal{P} \text{ and } C_t>C_s,\nonumber\\ 
    &\qquad~~ u^{(t)}, v^{(t)} \in \{0,1\}^{C_t} \text{ and } q^{(t)} \in \{0,1\}^{C_t\times K_t\times K_t} ~~ \forall t \in [L] \qquad\quad~~ (iii-a)~~ u^{(t)}_j \leq v^{(t)}_j \leq u^{(t)}_j + v^{(s)}_j \quad  \forall j \leq C_{s} \nonumber\\
    &\qquad\qquad\qquad\qquad\qquad\qquad\qquad\qquad\qquad\qquad\qquad\qquad\qquad\qquad\qquad  \qquad\qquad~~ v^{(t)}_j = u^{(t)}_j \quad \forall j > C_{s}\nonumber\\
    &\qquad\qquad\qquad\qquad\qquad\qquad\qquad\qquad\qquad\qquad\qquad\qquad\qquad\qquad\qquad  ~(iii-b)~~ u^{(t)} \preceq v^{(t)}. \nonumber
\end{align}
\normalsize
Concretely, \Cref{eq:efjointopt} reduces to the optimization problem for sequential convolution networks when $u^{(t)}\!=\!v^{(t)}$, $~~q^{(t)}_{j,a,b}\!=\!u_j^{(t)}$, and $\mathcal{P}\!=\!\emptyset$ $\forall t,j,a,b$.

%

\vspace{1em}
\textbf{ A.3 Skip concatenation}

Skip concatenation, which is a crucial feature of the well-known DenseNet \citep{densenet}, requires different techniques from skip addition. Concretely, the skip concatenation of a layer pair ($p, q$) means the $p$-th layer's feature map is concatenated with the $q$-th layer's feature map before the $q\!+\!1$-th convolution. To handle the possible skip concatenations, we utilize the fact that when $(p,q)$ is the skip concatenation pair, $q\!+\!1$-th convolution operation on the $q$-th layer can be thought as separate convolution operations on the $p$-th layer and the $q$-th layer, respectively. In this regard, we first assume there are convolution operations between every pair of layers. Then, we define $W^{(p,q)}\in \reals^{C_p\times C_q\times K_{p,q}\times K_{p,q}}$ as the convolution weights between $p$-th layer and $q$-th layer where $p\!<\!q$. If there is no skip concatenation from $p$-th layer to $q\!-\!1$-th layer, we regard there is no convolution operation between $p$-th layer and $q$-th layer and set $W^{(p,q)}=0_{K_{p,q}, K_{p,q}}$. Also, we introduce the corresponding shape column activation variables, $q^{(p,q)} \in \{0,1\}^{C_q\times K_{p,q}\times K_{p,q}}$, for the convolution operation from $p$-th layer to $q$-th layer. Then, we extend the optimization problem for skip concatenation as 

\footnotesize
\begin{align}
    \label{eq:jointoptdense}
    &\maximize_{r^{(0:L)}, q^{(1:L,1:L)}} ~~\sum_{p=1}^L \sum_{q=1}^L \left\langle I^{(p,q)}, A^{(p,q)} \right\rangle \\
    &~~\mathrm{subject~to} \nonumber\\
    &\qquad \sum_{p=0}^L a_p \left\|r^{(p)}\right\|_1 + \sum_{p=1}^L \sum_{q=1}^L b_{p,q} \left\|A^{(p,q)}\right\|_1 \leq M \nonumber \\
    &\qquad r^{(q)}_j \leq \sum_{a,b} q^{(p,q)}_{j,a,b} ~~~\text{ and } ~~~q^{(p,q)}_{j,a,b} \leq r_j^{(q)} \quad \forall p,q,j,a,b \nonumber\\
    &\qquad A^{(p,q)} = r^{(p)} \otimes q^{(p,q)} \quad \forall p,q \nonumber\\
    &\qquad r^{(p)} \in \{0,1\}^{C_p}, ~~ q^{(p,q)} \in \{0,1\}^{C_q\times K_{p,q}\times K_{p,q}}\quad \forall p,q \in [L].\nonumber
\end{align}
\normalsize

\section{Standard QCQP form}

\setcounter{proposition}{2}
\begin{proposition}
    \Cref{eq:optseq} is a QCQP problem. 
\end{proposition}
\begin{proof}
    We define the \emph{importance of 2-D filter}, which is the sum of the importance of weights in the filter as $F^{(l)}\in \reals_+^{C_{l\!-\!1} \times C_l} \quad \forall l$. Concretely, $F^{(l)}_{i,j} = \sum_{a,b} I^{(l)}_{i,j,a,b}\quad \forall i,j$. 
    We wish to express the objective function and constraints in \Cref{eq:optseq} with respect to $r^{(0:L)}$. Note that $\left\|A^{(l)}\right\|_1 = K_l^2\left\|r^{(l\!-\!1)}\right\|_1\left\|r^{(l)}\right\|_1$ and $\left\langle I^{(l)}, A^{(l)} \right\rangle = {r^{(l\!-\!1)}}^\intercal F^{(l)} {r^{(l)}}$.
    To express \Cref{eq:optseq} in a standard QCQP form, we denote $\left[r^{(0)}, r^{(1)}, \ldots, r^{(L)}\right]$ as $\mathbf{r}\in \{0,1\}^N$ where $N=\sum_{l=0}^L C_l$. Standard QCQP form of \Cref{eq:optseq} is
    \begin{align*}
        &\maximize_{\mathbf{r}\in \{0,1\}^N} \frac{1}{2} \mathbf{r}^\intercal P_0 \mathbf{r} \nonumber \\
        &~\mathrm{subject~to}\nonumber\\
        &\qquad\quad \frac{1}{2} \mathbf{r}^\intercal P_1 \mathbf{r} +  q_1^\intercal \mathbf{r} \leq M,
    \end{align*}
    where 
    \scriptsize
    \begin{align*}
        &P_0 =
    \begin{pmatrix}
        0                          &F^{(1)}              & 0                      &\cdots   & 0                          &0\\
        {F^{(1)}}^\intercal        & 0                   & F^{(2)}                &\cdots   & 0                          &0\\
        0                          &{F^{(2)}}^{\intercal}& 0                      &\cdots   & 0                          &0\\
        \vdots                     & \vdots              & \vdots                 &\ddots   & \vdots                     &\vdots\\
        0                          & 0                   & 0                      &\cdots   & 0                          &F^{(L)}\\
        0                          & 0                   & 0                      &\cdots   &{F^{(L)}}^{\intercal}       &0
    \end{pmatrix},\\
        &P_1=
    \begin{pmatrix}
        0           &Q_1        & 0     &\cdots   & 0       &0\\
        Q_1         &0          & Q_2   &\cdots   & 0       &0\\
        0           &Q_2        & 0     &\cdots   & 0       &0\\
        \vdots      &\vdots     &\vdots &\ddots   & \vdots  &\vdots\\
        0           & 0         & 0     &\cdots   & 0       & Q_L \\
        0           & 0         & 0     &\cdots   & Q_L     &0
    \end{pmatrix}\\
    \end{align*}\footnote{$Q_l = b_l K_l^2 J_{C_l}, \quad \forall l \in [L]$} 
    \normalsize
    , and $q_1=[\underbrace{a_0, \ldots, a_0}_{C_0} , \underbrace{a_1, \ldots, a_1}_{C_1}, \cdots, \underbrace{a_L, \ldots,  a_L}_{C_L}]$.
\end{proof}

\begin{proposition}
    \Cref{eq:jointoptseq} is a QCQP problem.
\end{proposition}
\begin{proof}
    To prove \Cref{eq:jointoptseq} is a QCQP problem, we show that objective function $\sum_{l=1}^L \left\langle I^{(l)}, A^{(l)} \right\rangle$, and the constraint $\sum_{l=0}^{L} a_l \left\|r^{(l)}\right\|_1 + \sum_{l=1}^L b_l \left\|A^{(l)}\right\|_1$, are sum of quadratic and linear terms of $r^{(0:L)}$ and $q^{(1:L)}$. 
    Note that
    \begin{align*}
        &\left\langle I^{(l)}, A^{(l)} \right\rangle=\sum_{i=1}^{C_{l\!-\!1}}\sum_{j=1}^{C_l}\sum_{a=1}^{K_l} \sum_{b=1}^{K_l}  I^{(l)}_{i,j,a,b}r^{(l\!-\!1)}_i q^{(l)}_{j,a,b}\\
        &\left\|r^{(l)}\right\|_1 = \sum_{i=1}^{C_l} r^{(l)}_i \\
        &\left\|A^{(l)}\right\|_1 = \sum_{i=1}^{C_{l\!-\!1}} \sum_{j=1}^{C_l} \sum_{a=1}^{K_l} \sum_{b=1}^{K_l} r^{(l\!-\!1)}_i q^{(l)}_{j,a,b}.
    \end{align*}
    Clearly, the objective function and all the constraints in \Cref{eq:jointoptseq} can be expressed as the sum of quadratic and linear terms of $r^{(0:L)}$ and $q^{(1:L)}$. Therefore, \Cref{eq:jointoptseq} is a QCQP problem with discrete variables, $r^{(0:L)}$ and $q^{(1:L)}$.
\end{proof}

\begin{proposition}
    \Cref{eq:efjointopt} is a QCQP problem.
\end{proposition}

\begin{proof}
    To prove \Cref{eq:efjointopt} is a QCQP problem, we show that $\sum_{t=1}^L \left\langle I^{(t)}, A^{(t)} \right\rangle$ and $\sum_{t=0}^{L} a_t \left\|u^{(t)}\right\|_1 + \sum_{t\in T} a_t \left\|v^{(t)}\right\|_1 + \sum_{t=1}^L b_t \left\|A^{(t)}\right\|_1$ are sum of quadratic and linear terms of $u^{(0:L)}, v^{(0:L)}$ and $q^{(1:L)}$. 
Note that
\begin{align*}
    &\left\langle I^{(t)}, A^{(t)} \right\rangle = \sum_{i=1}^{C_{t\!-\!1}}\sum_{j=1}^{C_t}\sum_{a=1}^{K_t} \sum_{b=1}^{K_t}  I^{(t)}_{i,j,a,b} v^{(t\!-\!1)}_i q^{(t)}_{j,a,b}\\
    &\left\|u^{(t)}\right\|_1 = \sum_{i=1}^{C_t} u^{(t)}_i\\
    &\left\|v^{(t)}\right\|_1 = \sum_{i=1}^{C_t} v^{(t)}_i\\
    &\left\|A^{(t)}\right\|_1 = \sum_{i=1}^{C_{t\!-\!1}} \sum_{j=1}^{C_t} \sum_{a=1}^{K_t} \sum_{b=1}^{K_t} v^{(t\!-\!1)}_i q^{(t)}_{j,a,b}.
\end{align*}
    Clearly, the objective function and all the constraints in \Cref{eq:efjointopt} can expressed by the sum of quadratic and linear terms of $u^{(0:L)}$, $v^{(0:L)}$ and $q^{(1:L)}$. Therefore, \Cref{eq:efjointopt} is a QCQP problem with discrete variables, $u^{(0:L)}$, $v^{(0:L)}$, and $q^{(1:L)}$.
\end{proof}

\section{Pruning consistency}

Pruning operation that removes weights through output channel direction leads to \emph{inactive} weights during the pruning procedure and prevent the exact modeling of the hard resource constraints (FLOPs and network size). In previous channel pruning methods based on the greedy approach, the pruned network requires post-pruning procedures to eliminate the remaining inactive weights. However, our formulation guarantees the exclusion of inactive weights from the pruned network.

\subsection{Preliminary}
    We assume each pruning methods outputs a pruning mask $A^{(t)}\in \{0,1\}^{C_{t\!-\!1}\times C_t \times K_l \times K_l}$. Then, we denote the pruned weights as $\hat{W}^{(t)} = W^{(t)} \odot A^{(t)}$. In the pruned network with pruned weights $\hat{W}^{(1:L)}$, we denote the input feature map of the $t\!+\!1$-th convolution as $V^{(t)}\in \reals^{C_t \times H_t \times W_t}$. Also, we denote the output feature map of the $t$-th convolution as $U^{(t)}\in \reals^{C_t \times H_t \times W_t}$. To avoid notation clutter, we ignore batch normalization and nonlinear activation function in this section. Then, $U^{(t)} = g^{(t)}(V^{(t\!-\!1)}; \hat{W}^{(t)})$, where $g^{(t)} : \reals^{C_{t\!-\!1} \times H_{t\!-\!1} \times W_{t\!-\!1}} \to \reals^{C_t \times H_t \times W_t}$, represents the convolution operation. In particular, 
    \begin{align}
        U^{(t)}_j = \sum_{i=1}^{C_{t\!-\!1}}g^{(t)}_{i,j} \left(V_i^{(t\!-\!1)} ; \hat{W}^{(t)}_{i,j}\right),
        \label{eq:convop}
    \end{align} 
    where $g^{(t)}_{i,j}: \reals^{H_{t\!-\!1} \times W_{t\!-\!1}} \to \reals^{H_t \times W_t}$ is a 2-D convolution operation with $\hat{W}^{(t)}_{i,j}$. Also, for ResNet, we formulate the relationship between the output feature map of a layer and the input feature map of the subsequent layer as 
    \begin{align}
        \label{eq:fdsc}
    &(i)~~~ V^{(t)} = U^{(t)} \quad \forall t \notin T \\
    &(ii)~~  V^{(t)} = U^{(t)} + V^{(s)}  \quad \forall (s,t) \in \mathcal{P} \text{ and } C_t=C_s\nonumber\\
    &(iii)~~ \forall (s,t) \in \mathcal{P} \text{ and } C_t>C_s,\nonumber\\ 
    &\quad(iii-a)~~ V^{(t)}_j =  U^{(t)}_j + V^{(s)}_j \quad  \forall j \leq C_{s} \text{ and } \nonumber\\
    &\quad\quad\quad\quad\quad V^{(t)}_j = U^{(t)}_j \quad \forall j > C_{s}\nonumber\\
    &\quad(iii-b)~~ V^{(t)} =  U^{(t)} + \tilde{V} ^{(s)}\quad \text{where} \nonumber\\
    &\quad\quad\quad\quad\quad \text{$\tilde{V}^{(s)}$ is $V^{(s)}$ after $1\times 1$ convolution.} \nonumber
    \end{align}

\subsection{Inactive weights}
    Before we specify inactive weights, we first define two important terms (\emph{trivially zero} and \emph{meaningless}). A feature map is \emph{trivially zero} if the feature map is zero for any input, $V^{(0)}$. A feature map is \emph{meaningless} if the values in the feature map do not have any effect on the final layer output feature map, $U^{(L)}$. Concretely, we state the definitions of trivially zero and meaningless in a cascading fashion.

\paragraph{Trivially zero} We define trivially zero in the ascending order of $t$. Concretely, we define trivially zero in the following order $V^{(0)} \rightarrow U^{(1)} \rightarrow V^{(1)} \rightarrow U^{(2)} \rightarrow \cdots \rightarrow V^{(L\!-\!1)} \rightarrow  U^{(L)}$.
            \begin{enumerate}
                \item $V^{(0)}_j$ in the input feature map is not trivially zero for all $j$.
                \item $U^{(t)}_j$ is trivially zero if and only if $A^{(t)}_{i, j}=0_{K_t,K_t} $ or $V^{(t\!-\!1)}_i$ is trivially zero for all $i\in [C_{t\!-\!1}]$ due to \Cref{eq:convop}. 
                \item In case of $V^{(t)}_j$, we divide the cases according to \Cref{eq:fdsc}.\\
                    $(i)$ $V^{(t)}_j$ is trivially zero if and only if $U^{(t)}_j$ is trivially zero.\\
                    $(ii)$ $V^{(t)}_j$ is trivially zero if and only if $U^{(t)}_j$ is trivially zero and $V^{(s)}_j$ is trivially zero.\\
                    $(iii-a)$ If $j\leq C_s$, the condition is the same with $(ii)$. Otherwise, the condition is the same with $(i)$.\\
                    $(iii-b)$ We suppose the output feature map of the $1\times 1$ convolution, $\tilde{V}^{(s)}_j$, is not trivially zero. Therefore, $V^{(t)}_j$ is not trivially zero.
            \end{enumerate}
\paragraph{Meaningless} We define meaningless in descending order of $t$. Concretely, we define meaningless in the following order $U^{(L)} \rightarrow V^{(L\!-\!1)} \rightarrow U^{(L\!-\!1)} \rightarrow V^{(L\!-\!2)} \rightarrow \cdots \rightarrow U^{(1)} \rightarrow  V^{(0)}$.
            \begin{enumerate}
                \item $U^{(L)}_j$ in the final feature map is not meaningless for all $j$.
                \item $V^{(t)}_i$ is meaningless if $A^{(t\!+\!1)}_{i, j}=0_{K_{t\!+\!1},K_{t\!+\!1}} $ or $U^{(t\!+\!1)}_j$ is meaningless for all $j\in [C_{t\!+\!1}]$ due to \Cref{eq:convop}.
                \item $U^{(t)}_i$ is meaningless if and only if $V^{(t)}_i$ is meaningless. 
            \end{enumerate}
    We now move on define \emph{active weight} and \emph{inactive weight} in \Cref{def:active} with trivially zero and meaningless.
    \begin{definition}[Active weight, inactive weight]
        \label{def:active}
        A weight $W^{(t)}_{i,j,a,b}$ is an \textbf{inactive weight} if 1) the weight is pruned ($A^{(t)}_{i,j,a,b}=0$) or 2) the corresponding input channel feature map ($V^{(t\!-\!1)}_i$) is trivially zero or 3) the corresponding output channel feature map ($U^{(t)}_j$) is meaningless. 
        Conversely, a weight $W^{(t)}_{i,j,a,b}$ is an \textbf{active weight} if 1) the weight is not pruned ($A^{(t)}_{i,j,a,b}=1$) and 2) the corresponding input channel feature map ($V^{(t\!-\!1)}_i$) is not trivially zero and 3) the corresponding output channel feature map ($U^{(t)}_j$) is not meaningless. 
    \end{definition}

    Note that only active weights should account for computation of the resource usage and the sum of the importance of weights. In this next subsection, we show that the inactive weights are provably excluded from the network pruned with our formulation.

\subsection{Pruning consistency in our formulation} 

In our method, discrete variables $u^{(0:L)}$, $v^{(0:L)}$, and $q^{(1:L)}$ satisfy the constraints in \Cref{eq:efconstraint}. We assume at least one of channel activation is set for each layer. Concretely, $\left\|u^{(t)}\right\|_1 \geq 1 \text{ and }\left\|v^{(t)}\right\|_1 \geq 1 \quad \forall t$.

\begin{subequations}
    \begin{align}
        &\left\|u^{(t)}\right\|_1 \geq 1 \text{ and }\left\|v^{(t)}\right\|_1 \geq 1 \quad \forall t \label{eq:efconstraint1} \\
        &u^{(t)}_j \leq \sum_{a,b} q^{(t)}_{j,a,b}  ~~~\text{and}~~~ q^{(t)}_{j,a,b} \leq u_j^{(t)} \quad \forall t,j,a,b \label{eq:efconstraint2} \\
        &(i)~~  v^{(t)} = u^{(t)} \quad \forall t \notin T\nonumber\\
        &(ii)~~  u^{(t)} \preceq v^{(t)} \preceq u^{(t)} + v^{(s)} \quad \forall (s,t) \in \mathcal{P} \text{ and } C_t=C_s \label{eq:efconstraint3}\\
        &(iii)~~ \forall (s,t) \in \mathcal{P} \text{ and } C_t>C_s,\nonumber\\ 
        &\qquad(iii-a)~~ u^{(t)}_j \leq v^{(t)}_j \leq u^{(t)}_j + v^{(s)}_j \quad  \forall j \leq C_{s} \text{ and } \nonumber\\
        &\qquad \qquad \qquad v^{(t)}_j = u^{(t)}_j \quad \forall j > C_{s}\nonumber\\
        &\qquad(iii-b)~~ u^{(t)} \preceq v^{(t)} \nonumber 
    \end{align}
    \label{eq:efconstraint}
\end{subequations}

\renewcommand\labelenumi{\theenumi.}

\begin{lemma}
    \label{lem:left}
    For $l\in [L]$, if $v^{(l\!-\!1)}_j=1$, then $V^{(l\!-\!1)}_j$ is not trivially zero. 
\end{lemma}
\begin{proof}
    We prove by mathematical induction with respect to $l$.\\
    \begin{enumerate}
        \item When $l=1$, the statement is true since input data is not trivially zero.
        \item Suppose the statement is true for $l=1,\ldots, t$.
        \item For $j$ such that $v^{(t)}_j=1$, we can think of three possible cases according to \Cref{eq:efconstraint3}\\
            \\
            $(i)$ If $t\notin T$,  $u^{(t)}_j=v^{(t)}_j=1$. First, $\exists i, ~~v^{(t\!-\!1)}_i=1$ since $\|v^{(t\!-\!1)}\|_1 \geq 1$ from \Cref{eq:efconstraint1}. By the induction hypothesis, $V^{(t\!-\!1)}_i$ is not trivially zero. On the other hand, $\exists a,b \quad q^{(t)}_{j,a,b}=1$ since $u^{(t)}_j \leq \sum_{a,b} q^{(t)}_{j,a,b}$ from \Cref{eq:efconstraint2}. Then, $A^{(t)}_{i,j,a,b}=v^{(t\!-\!1)}_i q^{(t)}_{j,a,b}=1$. By the second condition of trivially zero $(2)$, $U^{(t)}_j$ is not trivially zero since $V^{(t\!-\!1)}_i$ is not trivially zero and $A^{(t)}_{i,j,a,b}=1$. Also, by the third definition of trivially zero $(3-(i))$, $V^{(t)}_j$ is not trivially zero since $U^{(t)}_i$ is not trivially zero.\\
            $(ii)$ If $\forall (s,t) \in \mathcal{P}$ and $C_t=C_s$, $u^{(t)}_j + v^{(s)}_j \geq v^{(t)}_j=1$. Then, $u^{(t)}_j=1 $ or  $v^{(s)}_j=1$. If  $u^{(t)}_j=1$, $~~U^{(t)}_j$ is not trivially zero as in $(i)$. If  $v^{(s)}_j=1$, $~~V^{(s)}_j$ is not trivially zero by the induction hypothesis. By the definition of trivially zero $(3-(ii))$, $V^{(t)}_j$ is not trivially zero since $V^{(s)}_j$ or $U^{(t)}_j$ is not trivially zero.\\
            $(iii)$ $\forall (s,t) \in \mathcal{P}$ and $C_t>C_s$, \\
            $(iii-a)$ If $j \leq C_s$, the proof is the same with $(ii)$. Otherwise, the proof is the same with $(i)$. \\
            $(iii-b)$ By the definition of trivially zero $(3-(iii-b))$, $V_j^{(t)}$ is not trivially zero.
            In every possible cases, $V^{(t)}_j$ is not trivially zero and the statement is true for $l=t\!+\!1$. 
    \end{enumerate}
    By mathematical induction, for $l\in [L]$, if $v^{(l\!-\!1)}_j=1$, then $V^{(l\!-\!1)}_j$ is not trivially zero. 
\end{proof}

\begin{lemma}
    \label{lem:right}
    For $l\in [L]$, if $u^{(l)}_i=1$, then $U^{(l)}_i$ is not meaningless. 
\end{lemma}

\begin{proof}
    We prove by mathematical induction with respect to $l$.\\
    \begin{enumerate}
        \item When $l=L$, the statement is true since $U^{(L)}$ is not meaning less. 
        \item Suppose the statement is true for $l=t\!+\!1,\ldots, L$.
        \item For $i$ such that $u^{(t)}_i=1$, $v^{(t)}_i =1$ since $v^{(t)}_i\geq u^{(t)}_i$. Then, $\exists j \quad u^{(t\!+\!1)}_j=1$ since $\|u^{(t\!+\!1)}\|_1 \geq 1$ from \Cref{eq:efconstraint1}.
            By the induction hypothesis, $U^{(t\!+\!1)}_j$ is not meaningless. On the other hand, $\exists a,b \quad q^{(t\!+\!1)}_{j,a,b}=1$ since $u^{(t\!+\!1)}_j \leq \sum_{a,b} q^{(t\!+\!1)}_{j,a,b}$ from \Cref{eq:efconstraint2}.
            Then, $A^{(t\!+\!1)}_{i,j,a,b} = v^{t}_i q^{(t\!+\!1)}_{j,a,b} = 1$. By the definition of meaningless $(2)$, $V^{(t)}_i$ is not meaningless. By the definition of meaningless $(3)$, $U^{(t)}_i$ is not meaningless. The statement is true for $l=t$.
    \end{enumerate}
    By mathematical induction, for $l\in [L]$, if $u^{(l)}_i=1$, then $U^{(l)}_i$ is not meaningless. 
\end{proof}

\setcounter{proposition}{1}
\begin{proposition}
\label{prop2}
Optimizing over the input and output channel activation variables $u^{(0:L)}, v^{(0:L)}$ and shape column activation variables $q^{(1:L)}$ under the constraints in \Cref{eq:efconstraint} prevents the existence of any inactive weights in the pruned network guaranteeing exact computation of 1) resource usage and 2) the sum of the importance of active weights in the pruned network.
\end{proposition}
\begin{proof}
    Weight $W^{(l)}_{i,j,a,b}$ is not pruned if $A^{(l)}_{i,j,a,b}=1$. If $A^{(l)}_{i,j,a,b}=1$, then $u^{(l)}_j = 1$ and $v^{(l\!-\!1)}_i=1$. Then, by \Cref{lem:left} and \Cref{lem:right}, $V^{(l\!-\!1)}_i$ is not trivially zero and $U^{(l)}_j$ is not meaningless. By \Cref{def:active}, the weight $W^{(l)}_{i,j,a,b}$ is active. All the remaining weights in the network pruned with our method are active, which guarantees the exact specification of resource usage and sum of the importance of active weights in \Cref{eq:efjointopt}.
\end{proof}

\setcounter{proposition}{0}
\begin{proposition}
\label{prop1}
Optimizing over the input and output channel activation variables $r^{(0:L)}$ and shape column activation variables $q^{(1:L)}$ under the constraints in \Cref{eq:jointoptseq} prevents the existence of any inactive weights in the pruned network guaranteeing exact computation of 1) resource usage and 2) the sum of the importance of active weights in the pruned network.
\end{proposition}
\begin{proof}
    \Cref{prop1} is the special case of \Cref{prop2} when $u^{(t)}\!=\!v^{(t)}~\left(\coloneqq r^{(t)}\right)$, $q^{(t)}_{j,a,b}\!=\!u_j^{(t)}$, and $\mathcal{P}\!=\!\emptyset~~\forall t,j,a,b$.
\end{proof}

\section{Selecting a model  for inference time estimation}

\begin{figure}[h!]
\center
\begin{subfigure}[b!]{0.31\columnwidth}
\begin{tikzpicture}
\begin{axis}[width=4.35cm, height=4.4cm, grid=major, scaled ticks = false, ylabel near ticks, tick pos = left, 
tick label style={font=\scriptsize}, 
ytick={0,0.1,0.2,0.3},
xtick={0,0.1,0.2,0.3},
label style={font=\scriptsize}, xlabel={Estimated time (ms)}, ylabel={Actual time (ms)}, xmin=0, xmax=0.35, ymin=0, ymax=0.35] 
\addplot+[no markers,name path=A,color=blue,domain=0:1] {0.9*\x};
\addplot+[no markers,name path=B,color=black,domain=0:1] {\x};
\addplot+[no markers,name path=B,color=blue,domain=0:1] {1.1*\x};
\addplot+[red, only marks, mark options={fill=red, scale=0.6, mark=x, solid}] table [x=est, y=inf, col sep=comma]{csv/inf_exps/sol3.csv};
\addplot[no markers,blue!30] fill between[of=A and B];
\end{axis}
\end{tikzpicture}
\caption{M1}
\end{subfigure}
\begin{subfigure}[b!]{0.31\columnwidth}
\begin{tikzpicture}
\begin{axis}[width=4.35cm, height=4.4cm, grid=major, scaled ticks = false, ylabel near ticks, tick pos = left, 
tick label style={font=\scriptsize}, 
ytick={0,0.1,0.2,0.3},
xtick={0,0.1,0.2,0.3},
label style={font=\scriptsize}, xlabel={Estimated time (ms)}, ylabel={Actual time (ms)}, xmin=0, xmax=0.35, ymin=0, ymax=0.35] 
\addplot+[no markers,name path=A,color=blue,domain=0:1] {0.9*\x};
\addplot+[no markers,name path=B,color=black,domain=0:1] {\x};
\addplot+[no markers,name path=B,color=blue,domain=0:1] {1.1*\x};
\addplot+[red, only marks, mark options={fill=red, scale=0.6, mark=x, solid}] table [x=est, y=inf, col sep=comma]{csv/inf_exps/sol2.csv};
\addplot[no markers,blue!30] fill between[of=A and B];
\end{axis}
\end{tikzpicture}
\caption{M2}
\end{subfigure}
\begin{subfigure}[b!]{0.31\columnwidth}
\begin{tikzpicture}
\begin{axis}[width=4.35cm, height=4.4cm, grid=major, scaled ticks = false, ylabel near ticks, tick pos = left, 
tick label style={font=\scriptsize}, 
ytick={0,0.1,0.2,0.3},
xtick={0,0.1,0.2,0.3},
label style={font=\scriptsize}, xlabel={Estimated time (ms)}, ylabel={Actual time (ms)}, xmin=0, xmax=0.35, ymin=0, ymax=0.35] 
\addplot+[no markers,name path=A,color=blue,domain=0:1] {0.9*\x};
\addplot+[no markers,name path=B,color=black,domain=0:1] {\x};
\addplot+[no markers,name path=B,color=blue,domain=0:1] {1.1*\x};
\addplot+[red, only marks, mark options={fill=red, scale=0.6, mark=x, solid}] table [x=est, y=inf, col sep=comma]{csv/inf_exps/sol5.csv};
\addplot[no markers,blue!30] fill between[of=A and B];
\end{axis}
\end{tikzpicture}
\caption{M3}
\end{subfigure}
\begin{subfigure}[b!]{0.31\columnwidth}
\begin{tikzpicture}
\begin{axis}[width=4.35cm, height=4.4cm, grid=major, scaled ticks = false, ylabel near ticks, tick pos = left, 
tick label style={font=\scriptsize}, 
ytick={0,0.1,0.2,0.3},
xtick={0,0.1,0.2,0.3},
label style={font=\scriptsize}, xlabel={Estimated time (ms)}, ylabel={Actual time (ms)}, xmin=0, xmax=0.35, ymin=0, ymax=0.35] 
\addplot+[no markers,name path=A,color=blue,domain=0:1] {0.9*\x};
\addplot+[no markers,name path=B,color=black,domain=0:1] {\x};
\addplot+[no markers,name path=B,color=blue,domain=0:1] {1.1*\x};
\addplot+[red, only marks, mark options={fill=red, scale=0.6, mark=x, solid}] table [x=est, y=inf, col sep=comma]{csv/inf_exps/sol4.csv};
\addplot[no markers,blue!30] fill between[of=A and B];
\end{axis}
\end{tikzpicture}
\caption{M4}
\end{subfigure}
\begin{subfigure}[b!]{0.31\columnwidth}
\begin{tikzpicture}
\begin{axis}[width=4.35cm, height=4.4cm, grid=major, scaled ticks = false, ylabel near ticks, tick pos = left, 
tick label style={font=\scriptsize}, 
ytick={0,0.1,0.2,0.3},
xtick={0,0.1,0.2,0.3},
label style={font=\scriptsize}, xlabel={Estimated time (ms)}, ylabel={Actual time (ms)}, xmin=0, xmax=0.35, ymin=0, ymax=0.35] 
\addplot+[no markers,name path=A,color=blue,domain=0:1] {0.9*\x};
\addplot+[no markers,name path=B,color=black,domain=0:1] {\x};
\addplot+[no markers,name path=B,color=blue,domain=0:1] {1.1*\x};
\addplot+[red, only marks, mark options={fill=red, scale=0.6, mark=x, solid}] table [x=est, y=inf, col sep=comma]{csv/inf_exps/sol1.csv};
\addplot[no markers,blue!30] fill between[of=A and B];
\end{axis}
\end{tikzpicture}
\caption{M5}
\end{subfigure}
\begin{subfigure}[b!]{0.31\columnwidth}
\begin{tikzpicture}
\begin{axis}[width=4.35cm, height=4.4cm, grid=major, scaled ticks = false, ylabel near ticks, tick pos = left, 
tick label style={font=\scriptsize}, 
ytick={0,0.1,0.2,0.3},
xtick={0,0.1,0.2,0.3},
label style={font=\scriptsize}, xlabel={Estimated time (ms)}, ylabel={Actual time (ms)}, xmin=0, xmax=0.35, ymin=0, ymax=0.35] 
\addplot+[no markers,name path=A,color=blue,domain=0:1] {0.9*\x};
\addplot+[no markers,name path=B,color=black,domain=0:1] {\x};
\addplot+[no markers,name path=B,color=blue,domain=0:1] {1.1*\x};
\addplot+[red, only marks, mark options={fill=red, scale=0.6, mark=x, solid}] table [x=est, y=inf, col sep=comma]{csv/inf_exps/sol0.csv};
\addplot[no markers,blue!30] fill between[of=A and B];
\end{axis}
\end{tikzpicture}
\caption{M6}
\end{subfigure}
\caption{Estimated inference time vs. actual wall-clock inference time of convolution operations using each prediction models. The blue area indicates the error of the estimated value is under 10\% with respect to the actual value.}

\label{fig:infexp}
\end{figure}
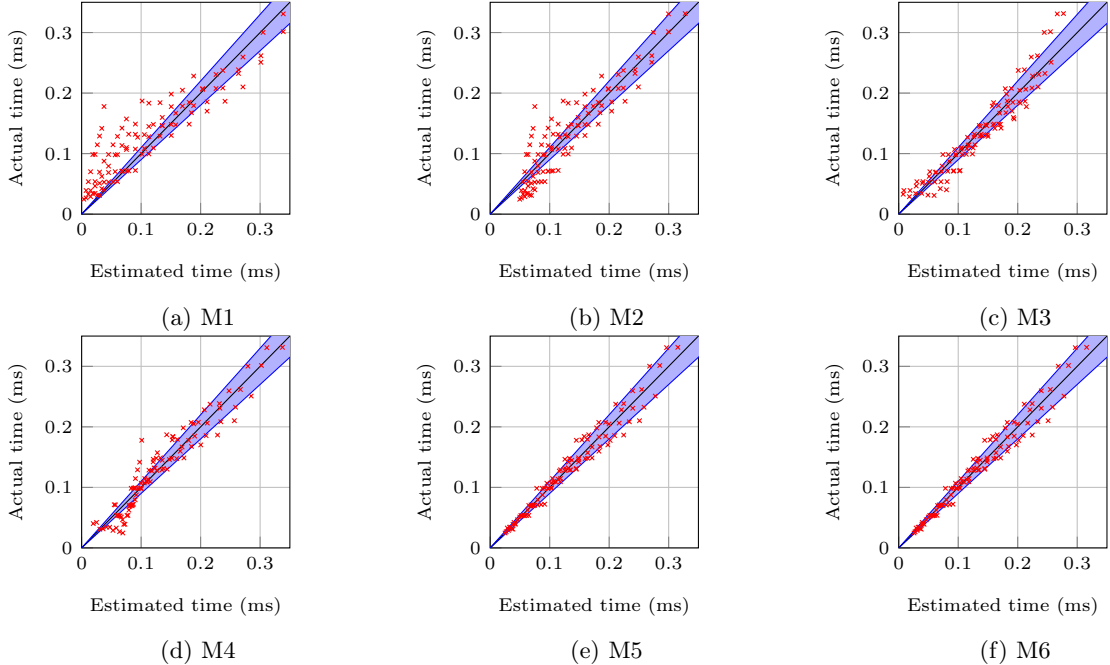

In this section, we aim to find a quadratic model that accurately predicts the inference time of convolution operations. We then analyze the coefficients of the proposed model and provide some intuitive meanings behind each term. Finally, we verify the proposed model is able to accurately predict the inference time of the whole pruned network as well, using ResNet-50 as the baseline network. We use a machine with Intel Xeon E5-2650 CPU and Titan XP GPU as default. In \Cref{subsec:inference_resnet}, we also experiment on another machine with Xeon Gold 5220R CPU and Geforce RTX 2080 Ti GPU.

Before we start, we define some notations to describe the convolution operation. We denote the kernel size, the stride, and the padding of a convolution operation as $k$, $s$ and $p$, respectively. Also, we denote the number of the input channels and the output channels as $n_\text{in}$ and $n_\text{out}$. Finally, we denote the shape of the input and the output feature map as $(n_\text{in}, H_\text{in}, W_\text{in})$ and $(n_\text{out}, H_\text{out}, W_\text{out})$, respectively.

\subsection{Model selection}
\label{subsec:model_selection}

    We first build a candidate set of quadratic models that estimate the inference time of a convolution operation and denote the models from M1 to M6, as described in \Cref{tab:analinfer}. Then, we measure the actual wall-clock inference times of the convolution operations without bias terms, where 
    $(H_\text{in},W_\text{in})\!=\!(56,56)$, $k\!=\!3$, $s\!=\!1$, $p\!=\!1$, and $(n_\text{in},n_\text{out})\in\{(16x,16y)\!\mid\! x,y\in[10]\}$. Concretely, the inference time samples can be represented as $\{(n_\text{in}^{(i)}, n_\text{out}^{(i)}, \text{WallClock}^{(i)} )\}_{i=1}^{100}$. Using these samples as the dataset, we find the best coefficients for each candidate models via least square regression. For example, in the case of M6, we solve the following optimization problem to find $\alpha^*$, $\beta^*$ , $\gamma^*$, and $\delta^*$ that best fits the inference time samples:
    \small
    \begin{align*}
        \min_{\alpha, \beta, \gamma, \delta} \sum_{i=1}^{100} \left( \text{WallClock}^{(i)} - \alpha - \beta n_\text{in}^{(i)} - \gamma n_\text{out}^{(i)} - \delta n_\text{in}^{(i)} n_\text{out}^{(i)} \right)^2.
    \end{align*}
	\normalsize
     \Cref{fig:infexp} shows the estimated inference time versus the wall-clock inference time when using each quadratic model with its best coefficients. We observe that M5 and M6 show the most successful estimation performance. More concretely, the mean percent error (MPE) and the $R^2$ value in \Cref{tab:analinfer} verifies that M5 and M6 are the most accurate models. That said, since M5 is a simpler model compared to M6 and the performance gap between the two models is negligible, we select M5 as our prediction model to estimate the inference time of convolution operations and analyze the coefficients of this model.

\begin{table}[h!]
    \caption{Description of the candidate quadratic models (M1-M6) for estimating the inference time of a convolution operation and
    their Mean Percent Error ($\%$) and $R^2$ values after regression over the inference time samples.}
\centering
\begin{adjustbox}{max width=1.0\columnwidth}
\begin{tabular}{clccc}
\toprule
    Notation & Model & Mean Percent Error ($\%$)  & $R^2$ \\
\midrule
    M1&$\delta n_{\text{in}}n_{\text{out}}$&28.9&0.687\\
    M2&$\alpha + \delta n_{\text{in}}n_{\text{out}}$&26.8&0.846\\
    M3&$\alpha + \beta n_{\text{in}} + \gamma n_{\text{out}}$&19.9 &0.912\\
    M4&$\alpha + \gamma n_{\text{out}} + \delta n_{\text{in}} n_{\text{out}}$&19.3 &0.918 \\
    M5&$\alpha + \beta n_{\text{in}} + \delta n_{\text{in}} n_{\text{out}}$&7.68 &\textbf{0.960}\\
    M6&$\alpha + \beta n_{\text{in}} + \gamma n_{\text{out}} + \delta n_{\text{in}} n_{\text{out}}$&\textbf{7.67} &\textbf{0.960}\\
\bottomrule
\end{tabular}
\end{adjustbox}
    \label{tab:analinfer}
\end{table}

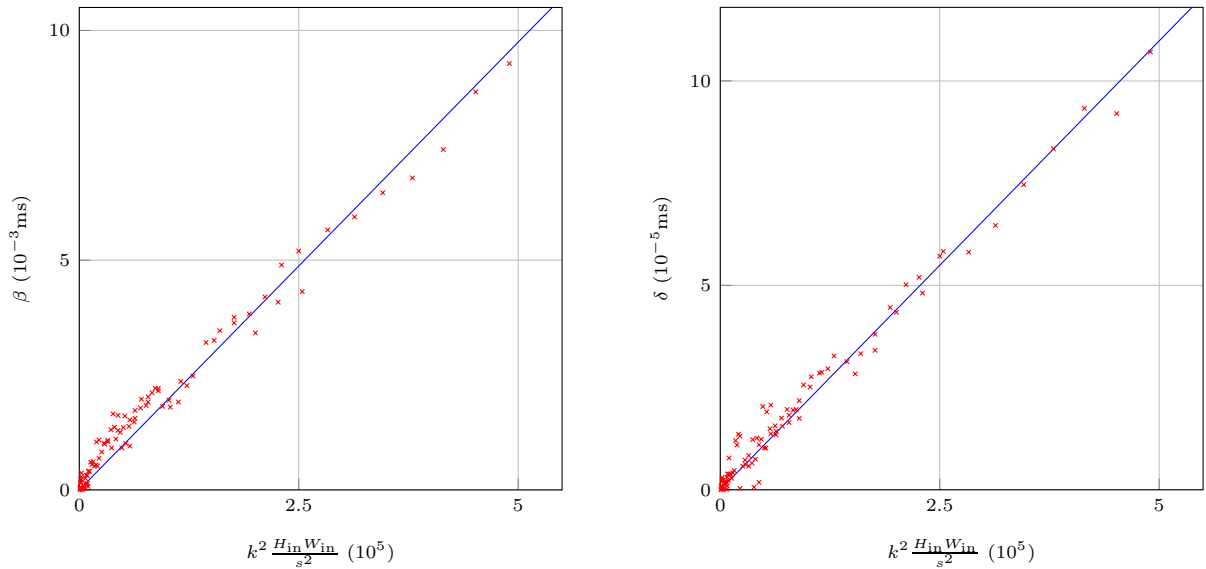
\begin{figure}[h]
\center
\begin{subfigure}[b!]{0.49\columnwidth}
\begin{tikzpicture}
\begin{axis}[width=8cm, height=8cm, grid=major, scaled ticks = false, ylabel near ticks, tick pos = left, 
tick label style={font=\scriptsize}, 
ytick={0,5,10},
xtick={0,2.5,5},
    label style={font=\scriptsize}, xlabel={$k^2\frac{H_\text{in} W_\text{in}}{s^2} ~(10^5)$}, ylabel={$\beta$~($10^{-3}$ms)}, xmin=0, xmax=5.5, ymin=0, ymax=10.5] 
\addplot+[no markers,color=blue,domain=0:5.5] {1.949*\x};
\addplot+[red, only marks, mark options={fill=red, scale=0.6, mark=x, solid}] table [x=cand, y=ans, col sep=comma]{supp_csv/inf_exps/val_sol_inv2.csv};
\end{axis}
\end{tikzpicture}
\end{subfigure}
\begin{subfigure}[b!]{0.49\columnwidth}
\begin{tikzpicture}
\begin{axis}[width=8cm, height=8cm, grid=major, scaled ticks = false, ylabel near ticks, tick pos = left, 
tick label style={font=\scriptsize}, 
ytick={0,5,10},
xtick={0,2.5,5},
    label style={font=\scriptsize}, xlabel={$k^2\frac{H_\text{in} W_\text{in}}{s^2} ~(10^5)$}, ylabel={$\delta~$($10^{-5}$ms)}, xmin=0, xmax=5.5, ymin=0, ymax=11.8] 
\addplot+[no markers,color=blue,domain=0:5.5] {2.196*\x};
\addplot+[red, only marks, mark options={fill=red, scale=0.6, mark=x, solid}] table [x=cand, y=ans, col sep=comma]{supp_csv/inf_exps/val_sol_inoutv2.csv};
\end{axis}
\end{tikzpicture}
\footnotesize
\end{subfigure}
\caption{Plot of $k^2\frac{H_\text{in} W_\text{in}}{s^2}$ vs. $\beta$ (left) and $\delta$ (right). The $R^2$ values without bias terms are $0.971$ (left) and $0.979$ (right). The blue lines are the linear lines without bias terms which best fit these points.}
\label{fig:infcoeff}
\end{figure}

\subsection{Analysis of the coefficients}
\label{subsec:coeff_analysis}

We conduct additional experiments to analyze the coefficients $\beta$ and $\delta$ of M5 by varying $H_\text{in}$, $W_\text{in}$, $k$, and $s$. We consider $96$ hyperparameter configurations of $(k, s, p, H_\text{in}, W_\text{in})$, sampled from $\{(2x-1, y, x-1, 52+4z, 52+4z)\mid (x,y,z)\in[4] \times [2] \times [12]\}$. For each sampled configuration, we measure the wall-clock inference times of $25$ different convolution operations with $(n_\text{in},n_\text{out})\in\{(16x,16y)\mid x, y \in [5]\}$ and fit the parameters $\alpha$, $\beta$, and $\delta$. As a result, we get 96 pairs of $\beta$ and $\delta$ values each corresponding to the 96 configurations. Examining the results, we observe a strong linear relationship between ($\beta, \delta$) and $k^2\frac{H_\text{in} W_\text{in}}{s^2}$, as illustrated in \Cref{fig:infcoeff}. Combining this with the fact that the FLOPs of a convolution operation is equal to $k^2\frac{H_\text{in} W_\text{in}}{s^2}n_\text{in} n_\text{out}$, we can hypothesize that $\delta n_\text{in} n_\text{out}$ in M5 represents the contribution of the computation cost (FLOPs) to the inference time. Also, the term $\beta n_\text{in}$ in M5 is proportional to $k^2\frac{H_\text{in} W_\text{in}}{s^2} n_\text{in}$. We assume that $\beta n_\text{in}$ represents the contribution of the memory access cost (MAC), as storing $\frac{H_\text{in}W_\text{in}}{s^2}$ patches of the input feature map, each of size $k^2 n_\text{in}$, accounts for the major part of the MAC. As more detailed calculation of the MAC may differ depending on the software and the hardware used, we leave further analysis as a future work.


\subsection{Inference time estimation on ResNet-50}
\label{subsec:inference_resnet}

We show our quadratic model can accurately predict the inference time of the whole ResNet-50 network as well as its convolution modules. Here, we also predict the inference time using M6. This is to take into account the batch normalization and the activation function that come after the convolution operation since these operations are only dependent on $n_{\text{out}}$. Considering these two functions with M6 enables a more accurate prediction of the wall-clock inference time of the whole network.

As a first step, we estimate the inference time of each convolution operation in ResNet-50. In particular, on M5, we estimate the inference time of only the convolution operation. On M6, we estimate the inference time of the convolution operation along with its subsequent operations (batch normalization and activation function). The results in \Cref{tab:layerexp} show that M5 and M6 successfully estimate the actual inference time on two different machines. Concretely, the average MPE value of M5 and M6 at a Titan XP machine are $6.48$\% and $7.98$\%, respectively. Also, the average MPE value of M5 and M6 at a RTX 2080 Ti machine are $11.66$\% and $10.66$\%, respectively.

\begin{figure}
\centering
    \captionsetup[subfigure]{justification=centering}
\begin{subfigure}[b!]{0.49\columnwidth}

\begin{tikzpicture}
\begin{axis}[width=8cm, height=8cm, grid=major, scaled ticks = false, ylabel near ticks, tick pos = left, 
tick label style={font=\normalsize}, 
ytick={1,1.1,1.2,1.3,1.4,1.5,1.6,1.7,1.8},
yticklabels={1,,1.2,,1.4,,1.6,,1.8},
xtick={1,1.1,1.2,1.3,1.4,1.5,1.6,1.7,1.8},
xticklabels={1,,1.2,,1.4,,1.6,,1.8},
label style={font=\normalsize}, xlabel={Estimated inference time (ms)}, ylabel={Wall-clock time (ms)}, xmin=1, xmax=1.85, ymin=1, ymax=1.85] 
\addplot+[no markers,name path=A,color=blue,domain=0:2] {1/0.9*\x};
\addplot+[no markers,name path=B,color=blue,domain=0:2] {1/1.1*\x};
\addplot+[red, only marks, mark options={fill=red, scale=0.6, mark=x, solid}] table [x=estimate, y=real, col sep=comma]{supp_csv/infer_scatter50_0.csv};
\addplot[no markers,blue!30] fill between[of=A and B];

\end{axis}
\end{tikzpicture}
\centering
\caption{
    ResNet50 and \\input ($224\times224$) with\\Titan XP GPU
}
\label{fiXPinfexptitan}
\end{subfigure}
\begin{subfigure}[b!]{0.49\columnwidth}

\begin{tikzpicture}
\begin{axis}[width=8cm, height=8cm, grid=major, scaled ticks = false, ylabel near ticks, tick pos = left, 
tick label style={font=\normalsize}, 
ytick={0.8,0.9,1,1.1,1.2,1.3,1.4},
yticklabels={0.8,,1,,1.2,,1.4,,1.6,,1.8},
xtick={0.8,0.9,1,1.1,1.2,1.3,1.4},
xticklabels={0.8,,1,,1.2,,1.4,,1.6,,1.8},
label style={font=\normalsize}, xlabel={Estimated inference time (ms)}, ylabel={Wall-clock time (ms)}, xmin=0.8, xmax=1.45, ymin=0.8, ymax=1.45] 
\addplot+[no markers,name path=A,color=blue,domain=0:2] {1/0.9*\x};
\addplot+[no markers,name path=B,color=blue,domain=0:2] {1/1.1*\x};
\addplot+[red, only marks, mark options={fill=red, scale=0.6, mark=x, solid}] table [x=estimate, y=real, col sep=comma]{supp_csv/scatter_total100.csv};
\addplot[no markers,blue!30] fill between[of=A and B];

\end{axis}
\end{tikzpicture}
\centering
\caption{
    ResNet50 and \\input ($224\times224$) with\\Geforce RTX2080 Ti GPU
}
\label{fig:infexpboth}
\end{subfigure}
\caption{
    Estimated inference time vs.\ actual wall-clock time of ResNet-50 architecture while varying the number of channels at a machine with Intel Xeon E5-2650 CPU and Titan XP GPU (a), and a machine with Xeon Gold 5220R CPU and Geforce RTX 2080 Ti GPU (b). The blue area indicates the error of the estimated value is under 10\% with respect to the actual wall-clock time.
}
\label{fig:infexp3}
\end{figure}
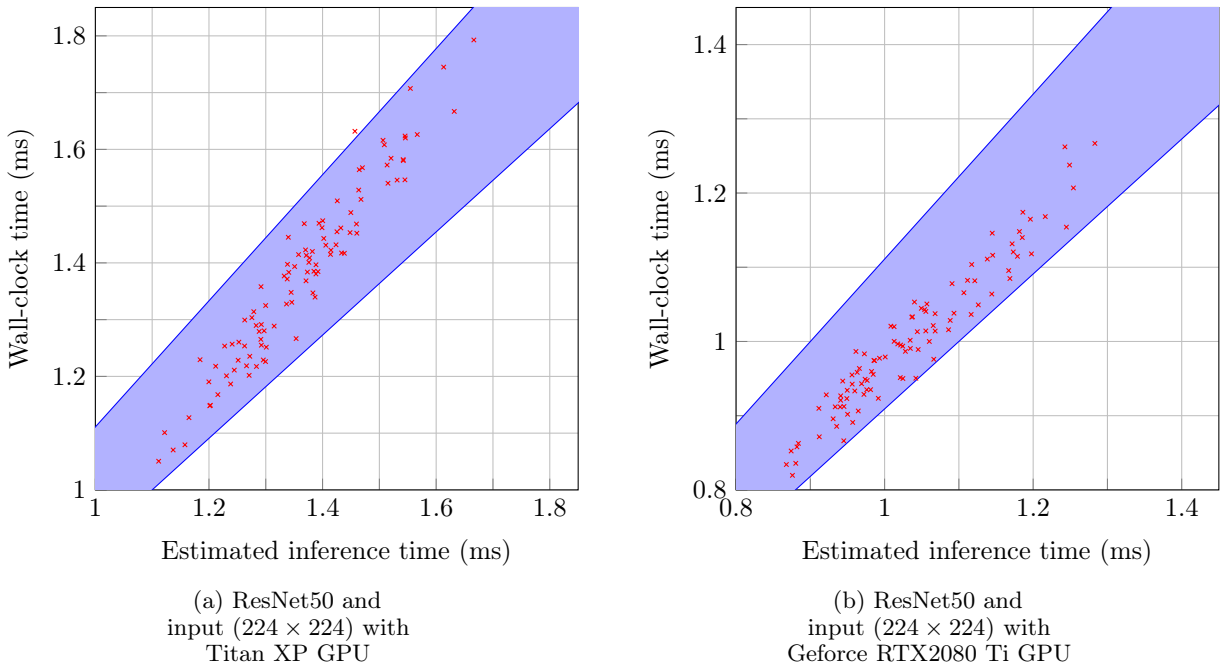

Next, we estimate the wall-clock inference time of the pruned ResNet-50 with M6 by summing all of the estimated inference time of the convolution modules, including the batch normalization and activation function. \Cref{fig:infexp3} shows the estimated inference time and the corresponding wall-clock inference time of ResNet-50 while varying the number of channels. Our proposed quadratic model M6 achieves 3.1\% and 3.5\% error rates on average when estimating the inference time of ResNet-50 from Titan XP and RTX 2080 Ti, respectively.

\begin{table}[h!]
    \caption{ Mean Percent Error (MPE) and $R^2$ values for estimating the inference time of each convolution layer in ResNet-50 architecture on a machine with Intel Xeon E5-2650 CPU and Titan XP GPU (`Titan XP'), and a machine with Xeon Gold 5220R CPU and Geforce RTX 2080 Ti GPU (`RTX 2080 Ti'). On M5, we estimate the inference time of only the convolution operation, while on M6 we also take into account the inference of the subsequent batch normalization and activation function (ReLU). `Avg' represents the averaged MPE and $R^2$ value of all convolutional layers in ResNet-50.}
\centering
    \begin{subtable}[b]{0.47\columnwidth}
        \caption{Convolution, M5}
        \centering
\begin{adjustbox}{max width=1.0\columnwidth}
\begin{tabular}{clcccccccccccc}
\toprule
&&&&&&&&\multicolumn{2}{c}{Titan XP}&\multicolumn{2}{c}{RTX 2080 Ti}\\
\cmidrule(r){9-10} \cmidrule(r){11-12}
&Layer&$H_\text{in}$&$W_\text{in}$&$n_\text{in}$&$n_\text{out}$&$k$&$s$&MPE (\%)&$R^2$&MPE (\%)&$R^2$\\
\cmidrule(r){1-10} \cmidrule(r){11-12}
conv1&1&224&224& 3& 64& 7& 2& 5.07& 0.85& 4.65& 0.10\\
\cmidrule(r){1-10} \cmidrule(r){11-12}
Block1&2&56&56& 64& 64& 1& 1& 1.21& 0.52& 0.95& 0.42\\
&3&56&56& 64& 64& 3& 1& 6.81& 0.97& 9.03& 0.93\\
&4&56&56& 64& 256& 1& 1& 3.46& 0.40& 10.30& 0.73\\
&5&56&56& 256& 64& 1& 1& 12.86& 0.80& 12.84& 0.87\\
\cmidrule(r){1-10} \cmidrule(r){11-12}
Block2&11&56&56& 256& 128& 1& 1& 13.25& 0.83& 13.83& 0.90\\
&12&28&28& 128& 128& 3& 2& 6.93& 0.59& 7.49& 0.90\\
&13&28&28& 128& 512& 1& 1& 1.03& 0.33& 11.20& 0.56\\
&14&28&28& 512& 128& 1& 1& 7.60& 0.55& 14.19& 0.78\\
&15&28&28& 128& 128& 3& 1& 15.42& 0.87& 10.83& 0.86\\
\cmidrule(r){1-10} \cmidrule(r){11-12}
Block3&23&28&28& 512& 256& 1& 1& 11.57& 0.78& 11.80& 0.93\\
&24&14&14& 256& 256& 3& 2& 6.45& 0.95& 6.69& 0.82\\
&25&14&14& 256& 1024& 1& 1& 1.93& 0.51& 13.13& 0.63\\
&26&14&14& 1024& 256& 1& 1& 2.77& 0.49& 11.64& 0.86\\
&27&14&14& 256& 256& 3& 1& 7.93& 0.95& 17.63& 0.76\\
\cmidrule(r){1-10} \cmidrule(r){11-12}
Block4&41&14&14& 1024& 512& 1& 1& 13.17& 0.66& 12.75& 0.90\\
&42&7&7& 512& 512& 3& 2& 10.70& 0.97& 11.35& 0.91\\
&43&7&7& 512& 2048& 1& 1& 6.32& 0.58& 9.84& 0.57\\
&44&7&7& 2048& 512& 1& 1& 11.75& 0.83& 9.84& 0.72\\
&45&7&7& 512& 512& 3& 1& 8.72& 0.97& 11.64& 0.92\\
\cmidrule(r){1-10} \cmidrule(r){11-12}
Avg&&&&&&&&6.48&0.67&11.66&0.74\\
\bottomrule
\end{tabular}
\end{adjustbox}

\end{subtable}
\hfill
\begin{subtable}[b]{0.47\columnwidth}
\vspace{0.5cm}
    \caption{Convolution, batch normalization, and activation function, M6}
\begin{adjustbox}{max width=1.0\columnwidth}
\begin{tabular}{clcccccccccccc}
\toprule
&&&&&&&&\multicolumn{2}{c}{Titan XP}&\multicolumn{2}{c}{RTX 2080 Ti}\\
\cmidrule(r){9-10} \cmidrule(r){11-12}
&Layer&$H_\text{in}$&$W_\text{in}$&$n_\text{in}$&$n_\text{out}$&$k$&$s$&MPE (\%)&$R^2$&MPE (\%)&$R^2$\\
\cmidrule(r){1-10} \cmidrule(r){11-12}
conv1&1&224&224& 3& 64& 7& 2& 4.79& 0.79& 10.20& 0.06\\
\cmidrule(r){1-10} \cmidrule(r){11-12}
Block1&2&56&56& 64& 64& 1& 1& 0.61& 0.33& 0.36& 0.60\\
&3&56&56& 64& 64& 3& 1& 5.06& 0.97& 4.99& 0.96\\
&4&56&56& 64& 256& 1& 1& 10.44& 0.84& 14.67& 0.91\\
&5&56&56& 256& 64& 1& 1& 10.30& 0.91& 10.00& 0.95\\
\cmidrule(r){1-10} \cmidrule(r){11-12}
Block2&11&56&56& 256& 128& 1& 1& 9.67& 0.92& 9.59& 0.95\\
&12&28&28& 128& 128& 3& 2& 11.90& 0.65& 8.34& 0.90\\
&13&28&28& 128& 512& 1& 1& 5.87& 0.82& 9.83& 0.82\\
&14&28&28& 512& 128& 1& 1& 10.60& 0.67& 12.28& 0.88\\
&15&28&28& 128& 128& 3& 1& 14.35& 0.91& 10.28& 0.86\\
\cmidrule(r){1-10} \cmidrule(r){11-12}
Block3&23&28&28& 512& 256& 1& 1& 10.43& 0.89& 9.55& 0.95\\
&24&14&14& 256& 256& 3& 2& 5.84& 0.96& 9.46& 0.82\\
&25&14&14& 256& 1024& 1& 1& 6.27& 0.66& 11.24& 0.83\\
&26&14&14& 1024& 256& 1& 1& 6.89& 0.67& 10.24& 0.90\\
&27&14&14& 256& 256& 3& 1& 5.40& 0.98& 15.93& 0.76\\
\cmidrule(r){1-10} \cmidrule(r){11-12}
Block4&41&14&14& 1024& 512& 1& 1& 13.38& 0.78& 11.55& 0.91\\
&42&7&7& 512& 512& 3& 2& 10.21& 0.98& 14.79& 0.86\\
&43&7&7& 512& 2048& 1& 1& 9.72& 0.69& 9.45& 0.86\\
&44&7&7& 2048& 512& 1& 1& 10.07& 0.91& 7.23& 0.92\\
&45&7&7& 512& 512& 3& 1& 5.63& 0.98& 10.90& 0.93\\
\cmidrule(r){1-10} \cmidrule(r){11-12}
Avg&&&&&&&&7.98&0.81&10.66&0.85\\
\bottomrule
\end{tabular}
\end{adjustbox}
\end{subtable}
    \label{tab:layerexp}
\end{table}
\section{Coordinate descent style optimization}

In this section, we provide a block coordinate descent-style optimization algorithm for solving \Cref{eq:efjointopt} in \Cref{coordalg}. Note that \Cref{eq:efjointopt} is the generalized version of Equation (3) in Section 3.2.

\begin{algorithm}[h!]
   \caption{Succinct channel and spatial pruning optimization via QCQP}
\begin{algorithmic}
    \STATE \textbf{Input : } $B$, $M$, $\gamma$, $a_l, a'_l, b'_l, I^{(l)}, \quad \forall l$
    \STATE Initialize $u^{(0:L)} ,v^{(0:L)}$.
    \STATE $M\coloneqq M/\gamma$.
    \FOR {$n=1,\ldots,$MAXITER}
    \FOR {$i=0,\ldots,L-B+1$}
    \STATE $z = \left[u^{(i:i+B-1)}, v^{(i:i+B-1)}\right]$
    \STATE $\tilde{f}(z) = f(z;\text{rest of the variables in $u,v$ fixed})$.
    \STATE Optimize  $\tilde{f}(z)$ with respect to $z$ under the constraints in \Cref{eq:efjointopt}.
    \ENDFOR
    \ENDFOR
    \STATE $M\coloneqq \gamma M$
    \STATE Optimize \Cref{eq:efjointopt} with respect to $q^{(1:L)}$ while fixing $u^{(0:L)}$ and $v^{(0:L)}$.
    \STATE \textbf{Output : } $u^{(0:L)}, v^{(0:L)}, q^{(1:L)}$
\end{algorithmic}
    \label{coordalg}
\end{algorithm}

We first set all shape column activation variables to $(q^{(t)}_{j,a,b} = u_j^{(t)} ~ \forall t,j,a,b)$. Then, we optimize over the input and output channel activation variables $(u^{(0:L)}, ~v^{(0:L)})$ in a block coordinate descent fashion with the resource constraint $M/\gamma$ where $\gamma$ is the average spatial sparsity smaller than $1$. Then, we optimize over the shape column activation variables $q^{(1:L)}$, fixing the input and output channel activation variables. In all experiments using Algorithm 1, $\gamma$ is decreased from `$M/(\text{Resource requirment of the initial network})$' to $1.0$ with a step size of $0.1$. After the pruning procedure, we employ one round of finetuning on the pruned network. Note that in \Cref{coordalg}, we denote the objective function in \Cref{eq:efjointopt} as $f(\cdot)$ when the shape column activations are all forced to match the output channel activations $(q^{(t)}_{j,a,b} = u_j^{(t)} ~ \forall t,j,a,b)$. $z$ denotes the concatenated variables of input and output channel activation in the target block. 

We additionally conducted an experiment to check the CPLEX performance of \Cref{coordalg} compared to direct optimization on \Cref{eq:efjointopt}, which we denote as Algorithm 0. However, Algorithm 0 is not scalable even in ResNet-20 on CIFAR-10. Therefore we compare Algorithm 0 and 1 for the first eight layers of ResNet-56. We set $B=2$ and adjust $\gamma$ with a step size of $0.1$. Algorithm 1 succeeds in increasing the objective value to $100.16$ in $12$ minutes, while Algorithm 0 reaches $106.27$ in 1 hour. Also, Algorithm 1 requires only 3 hours and 4GB memory for pruning ResNet-50 on ImageNet.


\section{Implementation details and Additional experiment results}

\begin{table}[!ht]
    \caption{Pruned accuracy and accuracy drop from the baseline network at given pruning ratios on various ResNet architectures (ResNet-20,32,56) at CIFAR-10.} 
	\centering
	\begin{adjustbox}{max width=\columnwidth}
        \setlength\tabcolsep{4pt}
        \begin{tabular}{lccccc}
            \toprule 
            Method&IC&Baseline acc&Pruned acc$\uparrow$& Acc drop$\downarrow$&Pruning ratio(\%)$\uparrow$\\
            \midrule
            \multicolumn{3}{l}{Network: Resnet-20}\\
            \midrule
            FPGM \citep{fpgm}&F            &92.21 (0.18)  &91.26 (0.24)           &0.95           &54.0\\
            ours-c &F	                  &92.21 (0.18)  &91.26 (0.18)           &0.95           &\textbf{54.1}\\
            ours-cs    &F                  &92.21 (0.18)  &\textbf{92.02}(0.10)  &\textbf{0.19}  &54.0\\            
            \midrule
            SFP \citep{sfp}   &F          &92.20 (0.18)  &90.83 (0.31)           &1.37           &42.2\\ 
            FPGM \citep{fpgm} &F          &92.21 (0.18)  &91.72 (0.20)           &0.49           &42.2\\             
            ours-c 	          &F       &92.21 (0.18)  &92.27 (0.17). &-0.06		&\textbf{42.3}\\
            ours-cs              &F       &92.21 (0.18)  &\textbf{92.35} (0.10)  &\textbf{-0.14}  &42.2\\
            \midrule
            \midrule
            \multicolumn{3}{l}{Network: Resnet-32}\\
            \midrule
            FPGM \citep{fpgm} &F           &92.88 (0.86)  &91.96 (0.76)           &0.92   &\textbf{53.2}\\ 
            ours-c                 &F     &92.88 (0.86)  &92.22 (1.02)  &0.66   &\textbf{53.2}\\
            ours-cs                    &F &92.88 (0.86)  &\textbf{92.78} (0.97)                          &\textbf{0.10}  &\textbf{53.2}\\
            \midrule
            SFP \citep{sfp}            &F &92.63 (0.70)  &92.08 (0.08)          &0.55           &41.5\\ 
            FPGM \citep{fpgm}        &F   &92.88 (0.86)  &92.51 (0.90)          &0.37           &41.5\\
            ours-c                  &F    &92.88 (0.86)	 &92.42 (0.77)	        &0.46	        &\textbf{42.7}\\
            ours-cs                  &F   &92.88 (0.86)	 &\textbf{92.83} (0.83)	&\textbf{0.05}	&\textbf{42.7}\\ 
            \midrule
            \midrule
            \multicolumn{3}{l}{Network: Resnet-56}\\
            \midrule
            SFP \citep{sfp}      &F      &93.59 (0.58)  &92.26 (0.31)            &1.33   &52.6\\ 
            FPGM \citep{fpgm}    &F       &93.59 (0.58)  &93.49 (0.13)            &0.10   &52.6\\
            SCP \citep{kang2020operationaware}	&T*	&93.69		&93.23			 &0.46	&51.5\\			            
            ours-c            &F         &93.59 (0.58)  &93.37 (0.96)             &0.22      &\textbf{52.7}\\
            ours-cs               &F     &93.59 (0.58)  &\textbf{93.69} (0.69)	 &\textbf{-0.10}	&52.6\\
            \bottomrule
		\end{tabular}
	\end{adjustbox}
	\label{tab:cifar10mem}
\end{table}

\subsection{Implementation details}
For ResNet experiments, we mostly follow the implementation from FPGM \citep{fpgm}. We apply batch normalization and remove bias weight in every convolution layer. Zero padding and $1\times 1$ convolution are used as the downsampling technique in CIFAR-10 \citep{cifar} and  ImageNet \citep{imagenet}, respectively. 

For CIFAR-10 experiments on ResNet \citep{resnet} architectures, we finetune the pruned model from the pretrained network given in \citet{fpgm} and follow the protocol of \citet{fpgm} for fair comparison. We finetune the pruned network for 200 epochs with batch size 128 and initial learning rate 0.01. Then, we adjust the learning rate at 60, 120, and 160 epoch by multiplying 0.2 each time. We use SGD optimizer with momentum $0.9$, weight decay $5\times10^{-4}$, and Nesterov momentum. For CIFAR-10 experiments on the DenseNet-40 architecture, we finetune the pruned network for 300 epochs with batch size 128 and initial learning rate 0.1. We adjust the learning rate at 150 and 225 epoch by multiplying 0.1. Here, we also use SGD optimizer with momentum $0.9$, weight decay $5\times10^{-4}$, and Nesterov momentum. 

For ImageNet experiments on ResNet architectures, we follow the protocol of FPGM and start from the pretrained network provided by PyTorch \citep{pytorch}. We finetune the pruned network for 80 epochs on ImageNet with batch size 384 and the initial learning rate of 0.015. Then, we adjust the learning rate at 30 and 60 epoch by multiplying 0.1. Here, we use SGD optimizer with momentum $0.9$ and weight decay $10^{-4}$.

For ImageNet experiments on EfficientNet-B0 \citep{efficientnet}, we use RMSProp optimizer with weight decay $0.9$, momentum $0.9$, and weight decay $1\times10^{-5}$. We train the baseline network for $250$ epochs using batch size $256$ and initial learning rate $0.008$ that decays by $0.97$ every $2.4$ epochs with $3$ warmup epochs. Then, we prune the baseline network and train the pruned networks using the same schedule with the baseline network except for using the initial learning rate $0.0008$. For the regularization of both the baseline network and the pruned network, we use RandAugment \citep{randaugment}, color jitters of factor $0.4$, stochastic depth \citep{stochastic} with survival probability $0.8$, and dropout \citep{dropout} of rate $0.2$.

For ImageNet experiments on VGG-16 \citep{simonyan2015deep}, we follow the protocol of \citet{molchanov2016pruning} for network training. We start from the pretrained network from Pytorch. We finetune the pruned network using SGD optimizer with a constant learning rate $10^{-4}$ and batch size 32 for 5 epochs.

\subsection{CIFAR-10 network size constraint}

We conduct the experiments of another resource constraint, network size on ResNet architectures. In the experiment tables \Cref{tab:cifar10mem}, network size of the pruned network is computed according to the resource specifications in \Cref{eq:optseq,eq:jointoptseq}. `Pruning ratio' in the tables denotes the ratio of pruned weights among the total weights in baseline networks. Also, we ignore the extra memory overhead for storing the shape column activations due to its negligible size compared to the total network size.

\subsection{MobileNetV2}
\begin{table}[h]
\caption{Top1 pruned accuracy and accuracy drop from the baseline network at given FLOPs on MobileNetV2 architecture at ImageNet}
	\centering
	\begin{adjustbox}{max width=\columnwidth}
		\begin{tabular}{lccccc}
			\toprule 
            Method&Top1 Pruned Acc$\uparrow$& Top1 Acc drop$\downarrow$&FLOPs(\%)$\downarrow$\\ 
            \midrule
            NetAdapt \citep{yang2018netadapt}&70.9&0.9&70\\
            AMC \citep{he2019amc}&70.8&1.0&70\\
            MetaPruning \citep{metaprun} &\textbf{71.2} & \textbf{0.6} &69\\
            \midrule
            ours-c&70.8&1.0&\textbf{67}\\
            ours-cs&70.2&1.6&\textbf{67}\\
            ours-c (tuned)&71.0&0.8&\textbf{67}\\
            ours-cs (tuned)&70.9&0.9&\textbf{67}\\
            \bottomrule
\end{tabular}
\end{adjustbox}
\label{tab:mobilenetV22}
\end{table}
We additionally apply our pruning method on MobileNetV2 \citep{howard2017mobilenets}. We start from the pretrained network with accuracy $71.88$ provided by Pytorch. Then, we finetune the pruned network for $150$ epochs using SGD optimizer with weight decay $0.00004$, momentum $0.9$, and batch size $256$. For the training schedule, we apply a learning rate warm-up for the initial five epochs, which steps up from $0$ to $0.05$. Then, we use the cosine learning rate decay for the remaining epochs. 

We show our experiment results with three recent pruning baselines NetAdapt \citep{yang2018netadapt}, AMC \citep{he2019amc}, and MetaPruning \citep{metaprun} in \Cref{tab:mobilenetV22}. ‘(tuned)’ indicates that the normalizing factor, $\gamma_{l}$, is tuned with grid search. A fixed value is used otherwise. Our method shows performance competitive to the other baselines, NetAdapt, AMC and MetaPruning. However, we note that our method is much more efficient than NetAdapt, AMC, and MetaPruning since NetAdapt requires repetitive finetuning steps for the proposed networks, AMC requires repetitive trial and error steps to train DDPG \citep{DDPG} agent, and MetaPruning trains PruningNet of which network size is at least $30$ times bigger than that of original model. 

\subsection{FCN-32s for segmentation}

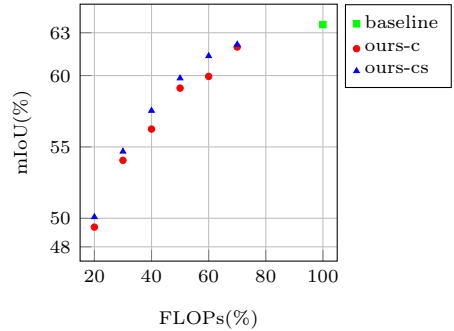
\begin{wrapfigure}[11]{r}{0.33\columnwidth} 
\vspace{-3em}
\center
\begin{tikzpicture}
\begin{axis}[width=5cm, height=5cm, only marks, grid=major, scaled ticks = false, ylabel near ticks, tick pos = left, 
tick label style={font=\scriptsize}, 
ytick={48,50,55,60,63}, 
xtick={20,40,60,80,100}, 
label style={font=\scriptsize}, xlabel={FLOPs(\%)}, ylabel={mIoU(\%)}, xmin=15, xmax=105, ymin=47, ymax=65,  legend style={legend columns=1, font=\scriptsize}, legend cell align={left}, legend pos=outer north east] 
\addlegendentry{baseline}
\addplot+[green, mark options={fill=green, scale=0.6, mark=square*, solid}] table  [x=f, y=mmm, col sep=comma]{supp_csv/fcn.csv};
\addlegendentry{ours-c}
\addplot+[red, mark options={fill=red, scale=0.6, mark=*, solid}]  table  [x=f, y=m, col sep=comma]{supp_csv/fcn.csv};
\addlegendentry{ours-cs}
\addplot+[blue, mark options={fill=blue, scale=0.6, mark=triangle*, solid}] table  [x=f, y=mm, col sep=comma]{supp_csv/fcn.csv};
\end{axis}
\end{tikzpicture}
\caption{The plot of mIoU (\%) versus FLOPs (\%) on FCN-32s.}
\label{fig:seg}
\end{wrapfigure}
We apply our pruning method on FCN-32s \citep{fcn} for segmentation on PASCAL Visual Object Classes Challenge 2011 dataset. Then, we evaluate the segmentation performance with a widely-used measure, mean Intersection over Union (mIoU). 

We use SGD optimizer with weight decay $0.0005$, momentum $0.99$, and batch size $1$. For the original network to be pruned, we train FCN-32s for $25$ epochs with a constant learning rate $8\times10^{-11}$. 
Then, we prune the original network of which mIoU $63.57$ (\%) and finetune the pruned network for $25$ epochs with a constant learning rate $8\times 10^{-11}$.

We show our experiment results in \Cref{fig:seg}. The pruned network reduces the FLOPs by 27\% with $0.15$ (\%) mIoU drop for `ours-c' and $0.09$ (\%) mIoU drop for `ours-cs'.

\section{Spatial pattern appearing in pruned network}

`ours-cs' discovers diverse spatial patterns in convolution weights, as illustrated in \Cref{fig:spat}. 
\newcommand\mask[9]{%
    \def\a{#1}
    \def\b{#2}
    \def\c{#3}
    \def\d{#4}
    \def\e{#5}
    \def\f{#6}
    \def\g{#7}
    \def\h{#8}
    \def\i{#9}
    \maskcontinued
    }
\newcommand\maskcontinued[2]{
    \def\x{#1}
    \def\y{#2}
    \draw[black] (1.4*\x,1.4*\y) rectangle (1.4*\x+1.2, 1.4*\y+1.2);
    \ifnum\a=1
    \draw[black,fill] (1.4*\x,1.4*\y) rectangle ++(0.4,0.4);
    \fi 
    \ifnum\b=1
    \draw[black,fill] (1.4*\x+0.4,1.4*\y) rectangle ++(0.4,0.4);
    \fi 
    \ifnum\c=1
    \draw[black,fill] (1.4*\x+0.8,1.4*\y) rectangle ++(0.4,0.4);
    \fi 
    \ifnum\d=1
    \draw[black,fill] (1.4*\x,1.4*\y+0.4) rectangle ++(0.4,0.4);
    \fi 
    \ifnum\e=1
    \draw[black,fill] (1.4*\x+0.4,1.4*\y+0.4) rectangle ++(0.4,0.4);
    \fi 
    \ifnum\f=1
    \draw[black,fill] (1.4*\x+0.8,1.4*\y+0.4) rectangle ++(0.4,0.4);
    \fi 
    \ifnum\g=1
    \draw[black,fill] (1.4*\x,1.4*\y+0.8) rectangle ++(0.4,0.4);
    \fi 
    \ifnum\h=1
    \draw[black,fill] (1.4*\x+0.4,1.4*\y+0.8) rectangle ++(0.4,0.4);
    \fi 
    \ifnum\i=1
    \draw[black,fill] (1.4*\x+0.8,1.4*\y+0.8) rectangle ++(0.4,0.4);
    \fi 
}
\begin{figure}[hbt!]
    \centering
    \begin{subfigure}[b]{0.31\columnwidth}
\resizebox{\columnwidth}{!}{
\begin{tikzpicture}]
\mask{1}{1}{1}{1}{1}{1}{1}{1}{0}{0}{0};
\mask{0}{1}{0}{1}{1}{0}{0}{0}{1}{0}{1};
\mask{0}{1}{0}{1}{1}{1}{0}{0}{0}{0}{2};
\mask{1}{0}{0}{1}{1}{1}{1}{0}{1}{0}{3};
\mask{0}{1}{1}{0}{1}{0}{0}{1}{0}{1}{0};
\mask{0}{1}{1}{0}{1}{1}{1}{1}{1}{1}{1};
\mask{0}{0}{0}{0}{0}{0}{0}{0}{0}{1}{2};
\mask{0}{1}{0}{1}{1}{1}{0}{1}{1}{1}{3};
\mask{0}{0}{0}{0}{0}{0}{0}{0}{0}{2}{0};
\mask{0}{0}{0}{0}{1}{0}{0}{0}{0}{2}{1};
\mask{0}{0}{0}{0}{0}{0}{0}{0}{0}{2}{2};
\mask{0}{0}{0}{0}{1}{0}{0}{0}{0}{2}{3};
\mask{0}{0}{0}{0}{0}{0}{0}{0}{0}{3}{0};
\mask{0}{0}{0}{1}{1}{1}{0}{0}{0}{3}{1};
\mask{0}{1}{0}{0}{1}{1}{0}{1}{0}{3}{2};
\mask{0}{0}{0}{0}{1}{0}{0}{0}{0}{3}{3};
\end{tikzpicture}}
\vspace{-1mm}
        \caption{$q^{(1)}$} 
\end{subfigure}
\hspace{5mm} 
\begin{subfigure}[b]{0.64\columnwidth}
\resizebox{\columnwidth}{!}{
\begin{tikzpicture}
\mask{1}{1}{1}{1}{0}{1}{1}{1}{1}{0}{0};
\mask{0}{1}{0}{0}{1}{0}{1}{1}{1}{0}{1};
\mask{1}{0}{1}{1}{0}{1}{1}{1}{1}{0}{2};
\mask{1}{1}{1}{0}{0}{1}{1}{1}{1}{0}{3};
\mask{0}{0}{1}{1}{0}{1}{0}{1}{1}{1}{0};
\mask{1}{1}{1}{0}{0}{0}{0}{1}{1}{1}{1};
\mask{1}{0}{1}{1}{0}{1}{1}{1}{1}{1}{2};
\mask{1}{1}{1}{0}{1}{1}{0}{0}{0}{1}{3};
\mask{1}{1}{1}{1}{0}{1}{1}{0}{1}{2}{0};
\mask{0}{1}{0}{0}{0}{1}{1}{1}{1}{2}{1};
\mask{0}{0}{1}{0}{0}{1}{1}{1}{1}{2}{2};
\mask{0}{1}{1}{0}{0}{1}{1}{1}{1}{2}{3};
\mask{1}{1}{0}{1}{0}{1}{1}{1}{1}{3}{0};
\mask{1}{0}{0}{0}{0}{1}{0}{1}{1}{3}{1};
\mask{0}{0}{0}{1}{0}{1}{1}{1}{1}{3}{2};
\mask{1}{1}{1}{0}{0}{0}{1}{1}{1}{3}{3};
\mask{1}{1}{0}{1}{0}{0}{1}{1}{1}{4}{0};
\mask{1}{1}{1}{1}{0}{1}{1}{0}{1}{4}{1};
\mask{1}{1}{0}{1}{1}{0}{0}{1}{1}{4}{2};
\mask{0}{1}{1}{1}{1}{1}{1}{1}{1}{4}{3};
\mask{1}{1}{0}{1}{0}{1}{1}{1}{1}{5}{0};
\mask{1}{1}{0}{0}{0}{0}{1}{1}{1}{5}{1};
\mask{1}{0}{1}{1}{1}{1}{1}{1}{1}{5}{2};
\mask{0}{0}{0}{1}{1}{0}{1}{1}{0}{5}{3};
\mask{1}{0}{1}{1}{1}{1}{1}{1}{1}{6}{0};
\mask{1}{1}{1}{1}{1}{0}{1}{1}{1}{6}{1};
\mask{1}{1}{0}{1}{0}{1}{0}{1}{1}{6}{2};
\mask{1}{1}{1}{0}{1}{1}{0}{1}{1}{6}{3};
\mask{1}{0}{1}{1}{0}{1}{1}{1}{1}{7}{0};
\mask{1}{1}{0}{1}{1}{1}{0}{1}{1}{7}{1};
\mask{0}{0}{1}{1}{0}{1}{1}{1}{1}{7}{2};
\mask{1}{1}{1}{0}{1}{1}{1}{1}{0}{7}{3};
\end{tikzpicture}}
\vspace{-1mm}
        \caption{$q^{(8)}$} 
\end{subfigure}
\caption{Visualization of shape column activations in the first convolution layer with $q^{(1)}\in \{0,1\}^{16\times 3\times 3}$ (left) and shape column activations in the eighth convolution layer with $q^{(8)}\in \{0,1\}^{32\times 3\times 3}$ (right) in ResNet-20 after pruned by `ours (c+s)'. Black area indicates that the shape column activation is set.}
\label{fig:spat}
\end{figure}
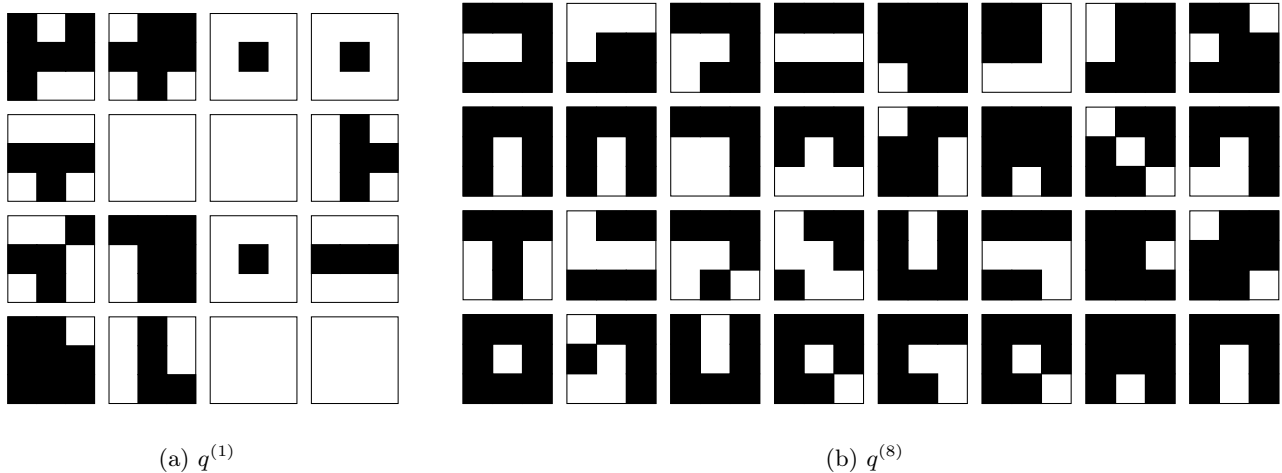

\section{Ablation study}
\begin{figure}[h!]
\centering
\begin{subfigure}[b!]{0.29\columnwidth}
\begin{tikzpicture}
\begin{axis}[width=5.3cm, height=5.3cm, only marks, grid=major, scaled ticks = false, ylabel near ticks, tick pos = left, 
tick label style={font=\normalsize}, 
ytick={7,8,9,10}, 
xtick={40,50,60,70,80}, 
label style={font=\normalsize}, xlabel={FLOPs(\%)}, ylabel={Error(\%)}, xmin=35, xmax=85, ymin=6.9, ymax=9.7] 
\addplot+[green, mark options={fill=green, scale=0.6, mark=*, solid}] table  [x=flops20, y=error20, col sep=comma]{csv/fpgmflops.csv};
\addplot+[blue, mark options={fill=blue, scale=0.6, mark=triangle*, solid}] table  [x=f20, y=flops_20, col sep=comma]{csv/ablation.csv};
\addplot+[red, mark options={fill=red, scale=0.9, mark=star, solid}] table  [x=f20, y=flops_20, col sep=comma]{csv/ablation_4d.csv};
\end{axis}
\end{tikzpicture}
\caption{ResNet-20}
\end{subfigure}
\begin{subfigure}[b!]{0.26\columnwidth}
\begin{tikzpicture}
\begin{axis}[width=5.3cm, height=5.3cm, only marks, grid=major, scaled ticks = false, ylabel near ticks, tick pos = left, 
tick label style={font=\normalsize}, 
ytick={7,8,9,10,11}, 
xtick={30,40,50,60,70,80}, 
label style={font=\normalsize}, xlabel={FLOPs(\%)}, xmin=25, xmax=85, ymin=6.5, ymax=11.5] 
\addplot+[green, mark options={fill=green, scale=0.6, mark=*, solid}] table  [x=flops32, y=error32, col sep=comma]{csv/fpgmflops.csv};
\addplot+[blue, mark options={fill=blue, scale=0.6, mark=triangle*, solid}] table  [x=f32,y=flops_32, col sep=comma]{csv/ablation.csv};
\addplot+[red, mark options={fill=red, scale=0.9, mark=star, solid}] table  [x=f32, y=flops_32, col sep=comma]{csv/ablation_4d.csv};
\end{axis}
\end{tikzpicture}
\caption{ResNet-32}
\end{subfigure}
\begin{subfigure}[b!]{0.36\columnwidth}
\begin{tikzpicture}
\begin{axis}[width=5.3cm, height=5.3cm, only marks, grid=major, scaled ticks = false, ylabel near ticks, tick pos = left, 
tick label style={font=\normalsize}, 
ytick={6,7,8,9,10}, 
xtick={30,40,50,60,70,80}, 
label style={font=\normalsize}, xlabel={FLOPs(\%)}, xmin=25, xmax=85, ymin=5.9, ymax=10.3, legend style={legend columns=1, font=\normalsize}, legend cell align={left}, legend pos=outer north east] 
\addlegendentry{FPGM}
\addplot+[green, mark options={fill=green, scale=0.6, mark=*, solid}] table  [x=flops56, y=error56, col sep=comma]{csv/fpgmflops.csv};
\addlegendentry{ours(c)}
\addplot+[blue, mark options={fill=blue, scale=0.6, mark=triangle*, solid}] table  [x=f56,y=flops_56, col sep=comma]{csv/ablation.csv};
\addlegendentry{ours(c+s)}
\addplot+[red, mark options={fill=red, scale=0.9, mark=star, solid}] table  [x=f56, y=flops_56, col sep=comma]{csv/ablation_4d.csv};
\end{axis}
\end{tikzpicture}
\caption{ResNet-56}
\end{subfigure}
\caption{The plots of classification error (\%) versus FLOPs (\%) on various ResNet architectures.}
\label{fig:flops_abl}
\end{figure}
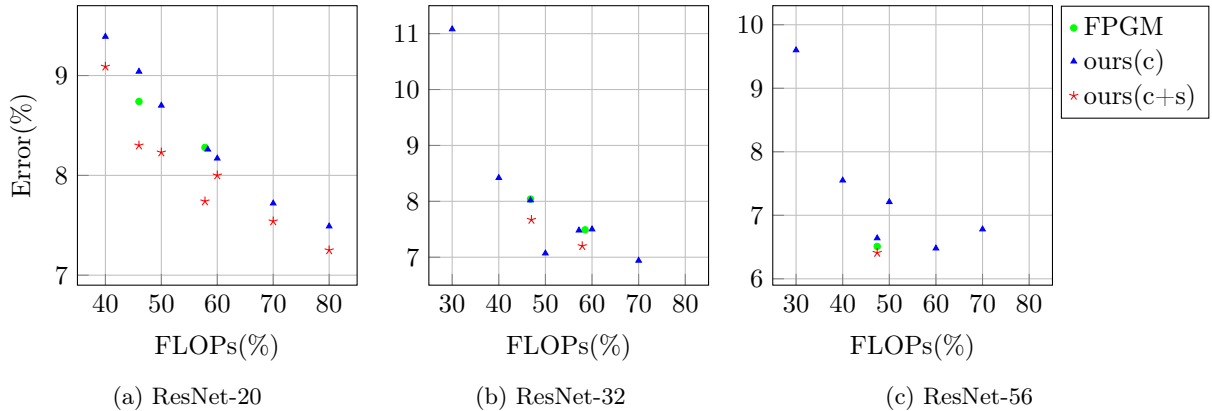
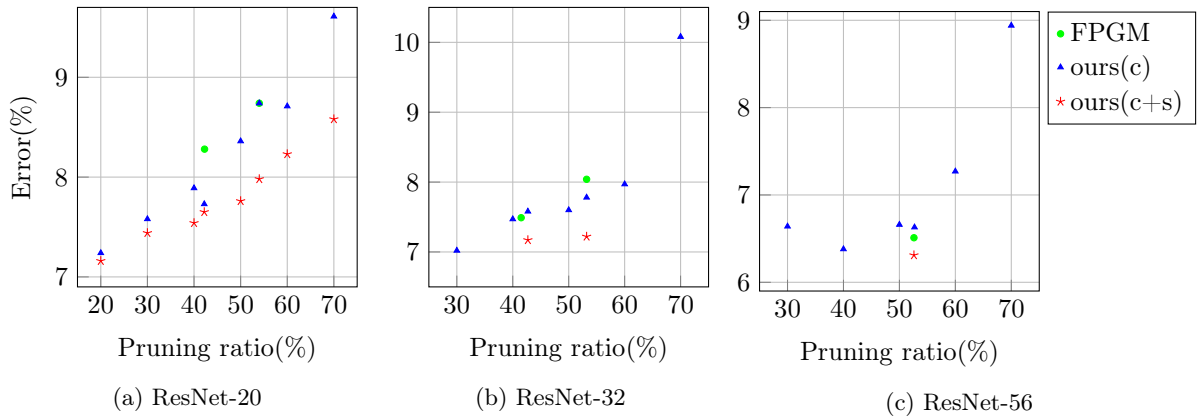
\begin{figure}[h!]
\centering
\begin{subfigure}[b!]{0.29\columnwidth}
\begin{tikzpicture}
\begin{axis}[width=5.3cm, height=5.3cm, only marks, grid=major, scaled ticks = false, ylabel near ticks, tick pos = left, 
tick label style={font=\normalsize}, 
ytick={7,8,9,10}, 
xtick={20,30,40,50,60,70}, 
label style={font=\normalsize}, xlabel={Pruning ratio(\%)}, ylabel={Error(\%)}, xmin=15, xmax=75, ymin=6.9, ymax=9.7, legend style={legend columns=1, font=\normalsize}, legend cell align={left}, legend pos=outer north east]
\addplot+[green, mark options={fill=green, scale=0.6, mark=*, solid}] table  [x=pr20, y=error20, col sep=comma]{csv/fpgmmem.csv};
\addplot+[blue, mark options={fill=blue, scale=0.6, mark=triangle*, solid}] table  [x=pr20, y=mem_20, col sep=comma]{csv/ablation.csv};
\addplot+[red, mark options={fill=red, scale=0.9, mark=star, solid}] table  [x=pr20, y=mem_20, col sep=comma]{csv/ablation_4d.csv};
\end{axis}
\end{tikzpicture}
\caption{ResNet-20}
\end{subfigure}
\begin{subfigure}[b!]{0.26\columnwidth}
\begin{tikzpicture}
\begin{axis}[width=5.3cm, height=5.3cm, only marks, grid=major, scaled ticks = false, ylabel near ticks, tick pos = left, 
tick label style={font=\normalsize}, 
ytick={6,7,8,9,10,11}, 
xtick={30,40,50,60,70}, 
label style={font=\normalsize}, xlabel={Pruning ratio(\%)}, xmin=25, xmax=75, ymin=6.5, ymax=10.5, legend style={legend columns=1, font=\normalsize}, legend cell align={left}, legend pos=outer north east]
\addplot+[green, mark options={fill=green, scale=0.6, mark=*, solid}] table  [x=pr32, y=error32, col sep=comma]{csv/fpgmmem.csv};
\addplot+[blue, mark options={fill=blue, scale=0.6, mark=triangle*, solid}] table  [x=pr32, y=mem_32, col sep=comma]{csv/ablation.csv};
\addplot+[red, mark options={fill=red, scale=0.9, mark=star, solid}] table  [x=pr32, y=mem_32, col sep=comma]{csv/ablation_4d.csv};
\end{axis}
\end{tikzpicture}
\caption{ResNet-32}
\end{subfigure}
\begin{subfigure}[b!]{0.36\columnwidth}
\begin{tikzpicture}
\begin{axis}[width=5.3cm, height=5.3cm, only marks, grid=major, scaled ticks = false, ylabel near ticks, tick pos = left, 
tick label style={font=\normalsize}, 
ytick={6,7,8,9,10}, 
xtick={30,40,50,60,70}, 
label style={font=\normalsize}, xlabel={Pruning ratio(\%)}, xmin=25, xmax=75, ymin=5.9, ymax=9.1, legend style={legend columns=1, font=\normalsize}, legend cell align={left}, legend pos=outer north east]
\addlegendentry{FPGM}
\addplot+[green, mark options={fill=green, scale=0.6, mark=*, solid}] table  [x=pr56, y=error56, col sep=comma]{csv/fpgmmem.csv};
\addlegendentry{ours(c)}
\addplot+[blue, mark options={fill=blue, scale=0.6, mark=triangle*, solid}] table  [x=pr56, y=mem_56, col sep=comma]{csv/ablation.csv};
\addlegendentry{ours(c+s)}
\addplot+[red, mark options={fill=red, scale=0.9, mark=star, solid}] table  [x=pr56, y=mem_56, col sep=comma]{csv/ablation_4d.csv};
\end{axis}
\end{tikzpicture}
\caption{ResNet-56}
\end{subfigure}
\caption{The plots of classification error (\%) versus pruning ratio (\%) on various ResNet architectures.}
\label{fig:size_abl}
\end{figure}
\begin{figure}[h!]
\centering
\begin{subfigure}[b!]{0.29\columnwidth}
\begin{tikzpicture}
\begin{axis}[width=5.3cm, height=5.3cm, only marks, grid=major, scaled ticks = false, ylabel near ticks, tick pos = left, 
tick label style={font=\normalsize}, 
ytick={0,20,40,60,80,100},
xtick={0,20,40,60,80,100},
label style={font=\normalsize}, xlabel={FLOPs(\%)}, ylabel={Pruning ratio(\%)}, xmin=0, xmax=100, ymin=0, ymax=100] 
\addplot+[blue, mark options={fill=blue, scale=0.6, mark=triangle*, solid}] table  [x=ours_f, y=ours_pr, col sep=comma]{csv/cmpr_fpgm_ours/resnet20.csv};
\addplot+[orange, mark options={fill=black, scale=0.9, mark=star, solid}] table  [x=f, y=pr, col sep=comma]{csv/ours_range2/resnet20.csv};
\addplot+[red, mark options={fill=red, scale=0.9, mark=*, solid}] table  [x=ours_f, y=ours_pr, col sep=comma]{csv/molchanov_range/resnet20.csv};
\end{axis}
\end{tikzpicture}
\caption{ResNet-20}
\end{subfigure}
\begin{subfigure}[b!]{0.26\columnwidth}
\begin{tikzpicture}
\begin{axis}[width=5.3cm, height=5.3cm, only marks, grid=major, scaled ticks = false, ylabel near ticks, tick pos = left, 
tick label style={font=\normalsize}, 
ytick={0,20,40,60,80,100},
xtick={0,20,40,60,80,100},
label style={font=\normalsize}, xlabel={FLOPs(\%)}, xmin=0, xmax=100, ymin=0, ymax=100] 
\addplot+[blue, mark options={fill=blue, scale=0.6, mark=triangle*, solid}] table  [x=ours_f, y=ours_pr, col sep=comma]{csv/cmpr_fpgm_ours/resnet32.csv};
\addplot+[red, mark options={fill=red, scale=0.9, mark=*, solid}] table  [x=ours_f, y=ours_pr, col sep=comma]{csv/molchanov_range/resnet32.csv};
\addplot+[orange, mark options={fill=black, scale=0.9, mark=star, solid}] table  [x=f, y=pr, col sep=comma]{csv/ours_range2/resnet32.csv};
\end{axis}
\end{tikzpicture}
\caption{ResNet-32}
\end{subfigure}
\begin{subfigure}[b!]{0.36\columnwidth}
\begin{tikzpicture}
\begin{axis}[width=5.3cm, height=5.3cm, only marks, grid=major, scaled ticks = false, ylabel near ticks, tick pos = left, 
tick label style={font=\normalsize}, 
ytick={0,20,40,60,80,100},
xtick={0,20,40,60,80,100},
label style={font=\normalsize}, xlabel={FLOPs(\%)}, xmin=0, xmax=100, ymin=0, ymax=100, legend style={legend columns=1, font=\normalsize}, legend cell align={left}, legend pos=outer north east]
\addlegendentry{\text{uniform}}
\addplot+[blue, mark options={fill=blue, scale=0.6, mark=triangle*, solid}] table  [x=ours_f, y=ours_pr, col sep=comma]{csv/cmpr_fpgm_ours/resnet56.csv};
\addlegendentry{\text{molchanov}}
\addplot+[red, mark options={fill=red, scale=0.9, mark=*, solid}] table  [x=ours_f, y=ours_pr, col sep=comma]{csv/molchanov_range/resnet56.csv};
\addlegendentry{\text{ours}}
\addplot+[orange, mark options={fill=black, scale=0.9, mark=star, solid}] table  [x=f, y=pr, col sep=comma]{csv/ours_range2/resnet56.csv};
\end{axis}
\end{tikzpicture}
\caption{ResNet-56}
\end{subfigure}
    \caption{The possible pairs of resource usage (FLOPs and pruning ratio) from three different pruning criteria : (1) `uniform' which greedily prunes channels layer-wise, (2) `molchanov' which greedily prunes the channels from all layers, and (3) `ours' which prunes via QCQP.}
\label{fig:range}
\end{figure}
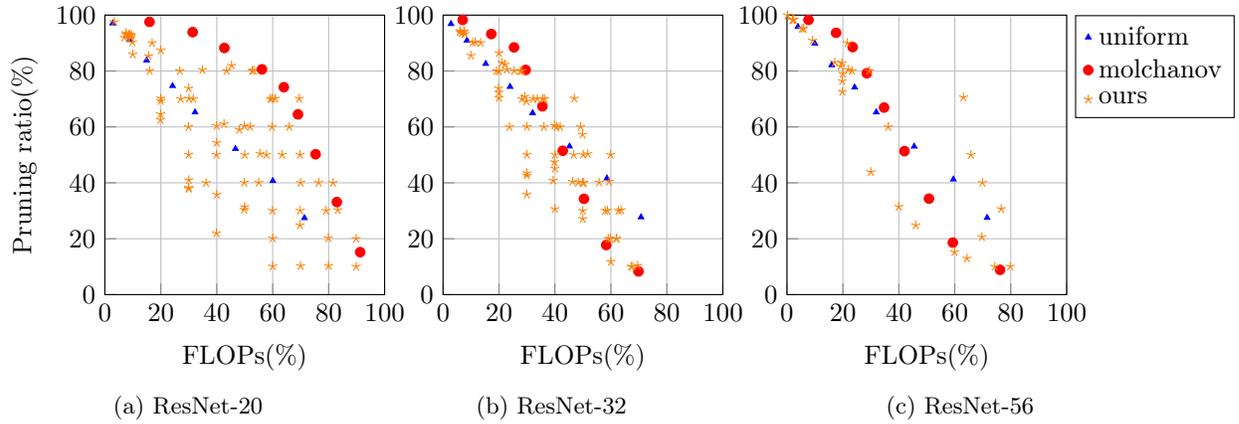

\subsection{Experiments on various FLOPs constraint}
In \Cref{fig:flops_abl}, we prune and finetune the ResNet architectures under various FLOPs constraints with `ours-c', `ours-cs', and FPGM. `ours-cs' outperforms `ours-c' and FPGM under almost all FLOPs constraints.
\subsection{Experiments on various network size constraint}
In \Cref{fig:size_abl}, we prune and finetune the ResNet architectures under various network size constraints with `ours-c', `ours-cs', and FPGM. `ours-cs' outperforms `ours-c' and FPGM on almost all network size constraints.

\subsection{Ablation study on the possible pairs of resource usage (FLOPs and pruning ratio)}
In the real-world, the resource constraint for network pruning may vary significantly in terms of how much each resource is available. In some cases, we may allow high FLOPs but strictly limit the network size, while in other cases, low FLOPs are much more important. However, when we prune the channels greedily, possible pairs of resource usages are limited and nonadjustable. In contrast, our method can target any resource budget pairs. \Cref{fig:range} shows the possible pairs (FLOPS and pruning ratio) from three different pruning methods: `uniform', which greedily prunes channels layer-wise, `molchanov', which greedily prunes the channels from all layers, and `ours', which prunes via QCQP. `ours' results in diverse pairs of target resources which cover the pairs of `uniform' and `molchanov'.

\end{document}